\documentclass[a4paper]{ar-1col} 
\usepackage[numbers]{natbib}
\usepackage{url}
\usepackage{amsmath}
\usepackage{amssymb}
\usepackage{bm}

\usepackage{algorithm}
\usepackage[noend]{algpseudocode}

\usepackage{subcaption}
\usepackage{graphicx}
\usepackage{xspace}
\usepackage{lscape}

\newcommand{\SE}[1]{\mathrm{SE}(#1)}
\newcommand{\pddl}[1]{\texttt{#1}}
\newcommand{\pddlkw}[1]{\textbf{\pddl{#1}}}

\newcommand{\figref}[1]{Figure \ref{#1}} 
\newcommand{\eref}[1]{Equation (\ref{#1})} 
\newcommand{\sref}[1]{Section \ref{#1}} 
\newcommand{\tref}[1]{Table \ref{#1}} 
\newcommand{\tamp}{{\sc tamp}\xspace}
\newcommand{\action}{action}

\algnewcommand{\LineComment}[1]{\State \(\triangleright\) #1}
\DeclareMathOperator*{\minimize}{minimize}

\newcommand{\field}[1]{\mathbb{#1}}
\newcommand{\R}{\field{R}} 
\newcommand{\B}{\{0, 1\}} 
\newcommand{\modefam}{\Sigma}

\usepackage{ifthen}
\usepackage{xcolor}
\newboolean{include-notes}
\setboolean{include-notes}{true}
\newcommand{\rhnote}[1]{\ifthenelse{\boolean{include-notes}}%
 {\textcolor{blue}{\textbf{RH: #1}}}{}}
\newcommand{\cgnote}[1]{\ifthenelse{\boolean{include-notes}}%
 {\textcolor{red}{\textbf{CG: #1}}}{}}
\newcommand{\lpknote}[1]{\ifthenelse{\boolean{include-notes}}%
 {\textcolor{olive}{\textbf{lpk: #1}}}{}}
\newcommand{\tlpnote}[1]{\ifthenelse{\boolean{include-notes}}%
 {\textcolor{purple}{\textbf{tlp: #1}}}{}}
 
\usepackage{colortbl}
\definecolor{Gray}{gray}{0.85}
\newcolumntype{g}{>{\columncolor{Gray}}c}
 
\usepackage[normalem]{ulem}
\useunder{\uline}{\ul}{}

\usepackage{listings}
\jname{Annu. Rev. Control Robot. Auton. Syst. 2021.}
\jvol{4}
\jyear{2021}

\begin{document}

\markboth{Garrett {\it et al.}}{Task and Motion Planning} 

\title{Integrated Task and Motion Planning} 

\author{Caelan Reed Garrett $^1$, 
Rohan Chitnis $^1$,
Rachel Holladay $^1$,
Beomjoon Kim $^1$,
Tom Silver $^1$,
Leslie Pack Kaelbling $^1$ and Tom\'as Lozano-P\'erez $^1$
\affil{$^1$CSAIL, MIT, Cambridge, USA, 02139; email: caelan@csail.mit.edu}
}

\begin{abstract}
The problem of planning for a robot that operates in environments containing a large number of objects, taking actions to move itself through the world as well as to change the state of the objects, is known as {\em task and motion planning} ({\sc tamp}).  
{\sc tamp} problems contain elements of discrete task planning, discrete-continuous mathematical programming, and continuous motion planning, and thus cannot be effectively addressed by any of these fields directly.
In this paper, we define a class of {\sc tamp} problems and survey algorithms for solving them, characterizing the solution methods in terms of their strategies for solving the continuous-space subproblems 
and their techniques for integrating the discrete and continuous components of the search.
\end{abstract}

\maketitle

\tableofcontents


\section{INTRODUCTION}

\begin{figure*}[ht]
    \vskip 0.1in
    \centering
    \includegraphics[width=\textwidth]{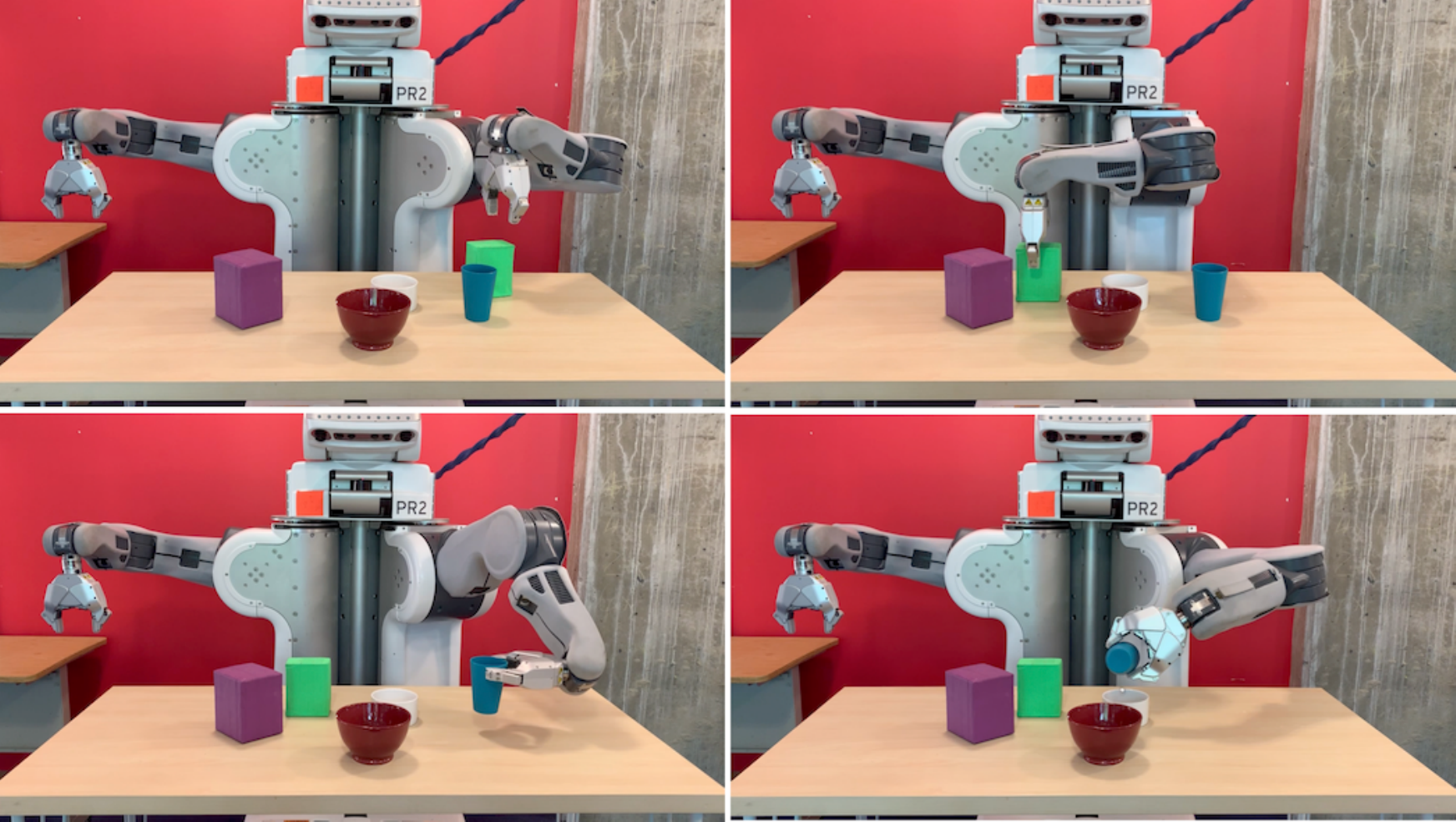}
    \caption{The specified goal is for the contents of the blue cup to end up in the white bowl.
    Because the green block obstructs reachable grasps for the blue cup, a {\sc tamp} algorithm automatically plans to relocate the green block before picking up the blue cup and pouring its contents into the white bowl. 
    From {\em left-to-right} and {\em top-to-bottom}: the robot picking up the green block, the robot placing the green block, the robot picking up the blue cup, and the robot pouring the blue cup's contents into the white bowl~\cite{wang2020learning}.}
    \label{fig:pr2} 
\end{figure*}

Robots are playing an increasingly important role in society, and their range of applications is rapidly expanding.  These applications have traditionally been in {\em structured} environments, such as factories, where the robot's interactions are limited and a behavior can be directly specified by a human.  However, many of the most exciting potential applications of robots are in highly \emph{unstructured} human environments such as homes, hospitals, or construction sites.  In these applications, the robot will generally be tasked with a specific goal, such as cooking and delivering a meal to an elderly resident, but the actions necessary to achieve the goal will vary enormously depending on the state of the environment.
For example, the robot might need to open cupboards and remove objects in order to 
retrieve a bowl that is necessary for
preparing the meal (\figref{fig:pr2}).  Directly specifying the full behavior policy for a robot operating in these unstructured environments is not practical because the required policy is too complex.

Since the earliest days of robotics, there has been an interest in automated {\em planning}, developing algorithms for deciding what sequence of commands the robot should execute in order to accomplish some goal~\cite{Fikes71,Nilsson84}.  The first class of planning problems that arises is to move the robot from one state to another without colliding with objects in the world.  This \emph{motion planning} problem was formulated by Lozano-P\'erez~\cite{LozanoPerez79} as a search for paths through the robot's {\em configuration space}, a space with dimensions representing the controllable joints of the robot, 
and has been the focus of a great deal of algorithmic development. 
The most effective methods are based on sampling~\cite{Kavraki96,lavalle2001randomized} or constrained optimization~\cite{ratliff2009chomp,schulman2014motion}. 

Collision-free robot motion is important but does not enable the robot to alter the world.  In order for the robot to, for example, move objects by picking them up and placing them, planning needs to consider a much larger space that encompasses the entire state of the world, which includes any objects the robot has grasped, the grasps it is using, and the poses of the other objects.  Conceptually, it makes sense to try to directly extend 
motion planning methods
to apply to entire world states, but this approach fails algorithmically.  
The entire world state, seen as a single kinematic system, is highly {\em under-actuated}, in the sense that from any configuration, most of the degrees of freedom cannot be changed at will.  The robot can only change the position of an object by moving over and touching it.

It is critical to understand the underlying topology of these spaces in order to plan in them.
Work by Alami {\it et al.}~\cite{Alami90ISRR,AlamiTwoProbs}, Branicky {\it et al.}~\cite{branicky2002nonlinear,branicky2006sampling}, and Hauser {\it et al.}~\cite{HauserLatombe,HauserIJRR11} observed that the configuration space of the world has important {\em modal} structure: depending on where the objects are placed and how they are grasped, the legal changes to the world are in a different {\em mode} or feasible submanifold of the full space.
Furthermore, it is only possible to change modes by moving to an intersection of the feasible space of the current mode and a new one, which is in general an even lower-dimensional subspace.  
For these reasons, planning is best viewed as a {\em hybrid} discrete-continuous search problem, of selecting a finite sequence of discrete mode types ({\it e.g.}, which objects to pick and place), continuous mode parameters (such as the poses and grasps of the movable objects),
and continuous motion paths within each mode to a configuration that is in the intersection with the subsequent mode.  


The artificial intelligence (AI) community has addressed problems of planning in very large discrete domains~\cite{ghallab2016automated}.  
Their techniques derive leverage from {\em factoring}, a source of combinatorial structure in planning problems.
Factoring is used to decompose the state space of the world into the Cartesian product of several subspaces, represented in terms of different state variables.  
Factoring enables compact representation of the {\em actions} that can be performed on a state:  these are generally described in terms of a small set of state variables that can be changed (while the others are held constant), as well as a condition on other variables that must be satisfied in order for the action to be executed.  
Furthermore, the AI planning community has developed a repertoire of very effective, domain-independent search algorithms that exploit this type of action representation~\cite{bonet2001planning,HoffmannN01,helmert2006fast}. 

Research in {\em task and motion planning} ({\sc tamp}) seeks to combine AI approaches to task planning and robotics approaches to motion planning.  A critical requirement for generality in approaches to {\sc tamp} actually lies {\em between} discrete ``high-level'' task planning 
and continuous ``low-level'' motion planning: 
an intermediate level of selecting the real-valued mode parameters, such as how to grasp and where to place an object, which govern legal continuous motions of the system.
This class of problems is computationally difficult in theory~\cite{deshpande2016tamp,vendittelli2015decidability} and requires algorithmic sophistication in practice.

\subsection{Example}
An essential component of {\sc tamp} problems is the interdependence of the motion-level and task-level aspects of the problem. 
Approaches that treat these independently, without considering their complex interplay, are unable to solve the general class of problems.  
Consider a problem in which the robot's goal is for a particular pot (named \pddl{A}) to be placed on one of the burners of the stove.   If the planner ignores the geometric aspects, it might select a high-level plan ``skeleton'' of the form:
\begin{equation*} 
    [\pddl{moveF}(q_0, \tau_1, q_1, p_0), \pddl{pick[A]}(q_1, p_0, g), 
    \pddl{moveH[A]}(g, q_1, \tau_2, q_2), \pddl{place[A]}(q_2, p_1, g)]
\end{equation*}
\noindent
where the \pddl{moveF} action involves robot movement when its hand is free, and the \pddl{moveH[A]} action involves robot movement when holding object \pddl{A}\footnote{See~\figref{fig:tamp-actions} for a complete definition of these actions.}.
This plan skeleton has free parameters involving robot configurations ($q_0, q_1, q_2$), a grasp pose ($g$), placement poses ($p_0, p_1$), and paths ($\tau_1, \tau_2$).  
The skeleton 
imposes constraints on the choices of those values
that will enable the plan to achieve the goal. Given this skeleton, it is now necessary to find values for all of these parameters that satisfy the constraints.   It may be that there is no satisfying set of values; for instance, a kettle could be occupying the target burner, preventing any safe placement of the pot.  In this case, a new skeleton is necessary: the robot will need to first move away the kettle and then place the pot on the stove.
This example demonstrates a change in the high-level plan that is necessitated by the low-level geometry.  


\subsection{Scope}
To keep the scope of this survey manageable, we limit the class of problems addressed, and discuss a variety of extensions in \sref{sec:extensions}.  In particular, we assume that: 1) actions are deterministic, 
2) the state of the world is completely known, 3) the robot and every object in the environment is a kinematic assembly of rigid bodies with known shapes, 
4) the robot is holonomic,
and 5) the goal is specified as a set of requirements on the final robot configuration, object poses, and possibly other state variables such as the cooked state of a dish.  
This class of problems encompasses a number of problems studied in the robotics community: {\em pick-and-place planning}~\cite{garrettIJRR2017}, {\em manipulation planning}~\cite{simeon2004manipulation}, {\em navigation among movable obstacles} ({\sc namo})~\cite{stilman2005navigation}, and {\em rearrangement planning}~\cite{KingRearrangementSpaces}.
An important related line of work uses
{\em linear temporal logic} (LTL) to provide high-level specifications for \tamp{} problems with temporally extended goals~\cite{belta2007symbolic,plaku2016motion}, but it is beyond the scope of this survey.

We begin with basic background in motion planning, multi-modal motion planning, and task planning (\sref{sec:background}).  
Next, we draw from components of these fields in order to formalize {\sc tamp} in a manner that allows for many existing approaches to be studied (\sref{sec:problem}).
We then describe a framework for understanding a broad class of {\sc tamp} algorithms in terms of combining (\sref{sec:combining}) a search over discrete plan structures with a search over continuous values satisfying constraints (\sref{sec:hcsp}) induced by the discrete structure.
We conclude with a short discussion of a rich array of extensions and generalizations of this basic problem class and the approaches to solve them (\sref{sec:extensions}).

\section{BACKGROUND}
\label{sec:background}
{\sc tamp} rests on foundations in robot motion planning (\sref{sec:motion}), multi-modal motion planning (\sref{sec:multi}), and AI task planning (\sref{sec:task}). In this section, we give a compact overview of each of these planning problem classes.

\subsection{Motion planning} \label{sec:motion}
The problem of planning motions for a robot with $d$ degrees of freedom can be framed as finding a trajectory for a point representing the robot's configuration through a $d$-dimensional configuration space.
More formally, a motion-planning problem is specified by a configuration space ${\cal Q} \subset \R^d$, a constraint $F: {\cal Q} \to \B$, an initial configuration $q_0 \in {\cal Q}$, and a goal set of configurations $Q_* \subseteq {\cal Q}$.
The {\em feasible} configuration space is a subset of ${\cal Q}$ that satisfies the constraint:  $Q_F = \{q \in {\cal Q} \mid F(q) = 1\}$.
The objective is to find a {\em continuous} path $\tau: [0, 1] \to {\cal Q}$ such that $\tau(0) = q_0$, $\tau(1) \in Q_*$, and $\forall \lambda \in [0, 1]\; \tau(\lambda) \in Q_F$.
The simplest motion-planning problems involve free-space motion, in which the robot simply needs to move through space without colliding.
Given a set of objects, defined by their shapes and poses in the world, the constraint $F(q)$ requires that the robot not collide with any object. 

Motion planning is {\sc pspace}-hard, but there are exact algorithms that leverage algebraic geometry to solve problems using only polynomial space (proving motion planning is {\sc pspace}-complete)~\cite{canny1988complexity}.
Despite this, the two most widely used approaches are {\em sampling-based} motion~planning~\cite{Kavraki96,lavalle2001randomized}
and {\em trajectory optimization}~\cite{ratliff2009chomp,schulman2014motion}.
Both classes of algorithms are useful in practice, but are not {\em complete} due to the fact that they cannot identify infeasible problems.
Many sampling-based motion planning algorithms can however, under some robustness conditions, be shown to be {\em probabilistically complete}, meaning that the probability that they will fail to find a solution, if one exists, converges to zero as the running time increases.
LaValle~\cite{Lavalle06} provides a comprehensive overview of motion planning algorithms.

\subsection{Multi-modal motion planning} \label{sec:multi}
{\em Multi-modal motion planning} ({\sc mmmp}) extends the problem space of planning to include changing the state of other objects in the world~\cite{HauserLatombe,hauser2010randomized,HauserIJRR11,barry2013hierarchical}.  To formalize {\sc mmmp} problems, we need to model changes in the kinematics of the system, extend motion planning to handle constraints beyond collision avoidance, and integrate these components. 

\subsubsection{Kinematic graphs}
\label{sec:kinematic graph}
One way to represent the geometric state of many environments is to encode the state variables collectively as a {\em kinematic graph}~\cite{Lavalle06}, which makes their dependencies explicit. 
In a kinematic graph, vertices represent bodies and the robot's controllable joints, and edges represent attachments.
Each edge has an associated relative transformation between the child body and parent body, which is a pose in $\SE{3}$. 
If each body is connected to at most one parent body and the graph is acyclic, this is a {\em kinematic tree}, 
for which the full state of the world can be derived from just the joint values of the robot $q$ through {\em forward kinematics}.  
The attachments can be of several kinds. The most straightforward is a rigid attachment, which models 
an object resting stably on a surface
or a robot grasping an object in a fixed grasp.  

\begin{figure}[h]
    \includegraphics[width=0.7\textwidth]{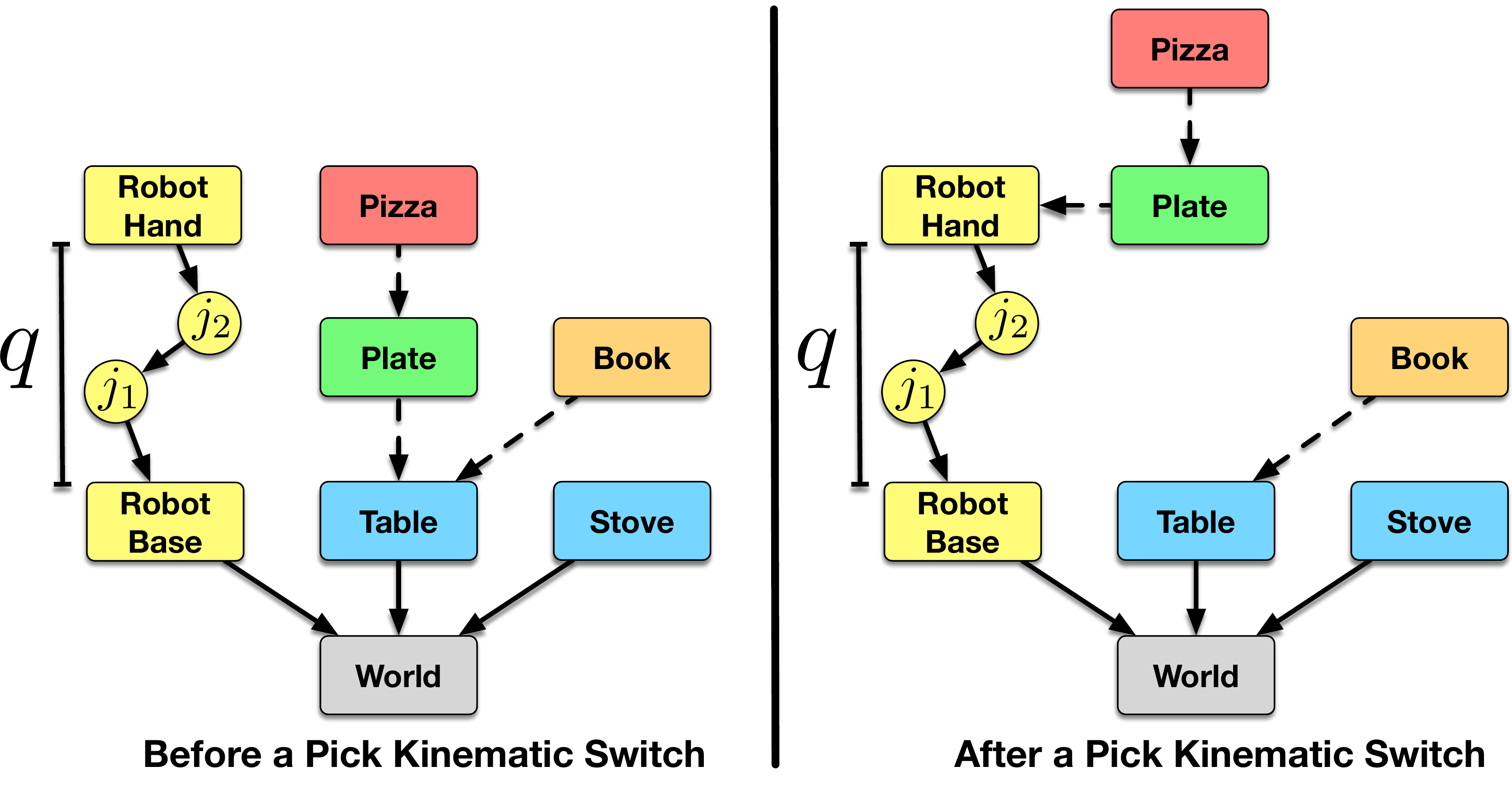}
    \caption{The change to a kinematic tree due to picking up the plate.  Square nodes are bodies, and round nodes are robot joints.  Lines encode attachments;  solid ones are fixed, and dotted ones may be changed.
    The robot has two joints ($j_1, j_2$), whose current state is given by configuration $q$. 
    }
    \label{fig:kinematic}
\end{figure}

\figref{fig:kinematic} represents a kinematic tree for an example kitchen environment that contains a robot manipulator with two joints ($j_1, j_2$), 
a fixed \pddl{Table} and \pddl{Stove}, and movable objects \pddl{Plate}, \pddl{Pizza}, and \pddl{Book}. 
Initially, \pddl{Pizza} rests on the \pddl{Plate}, which itself rests on the \pddl{Table}. 
When the robot picks up the \pddl{Plate}, it also transitively picks up \pddl{Pizza}.
This change in the kinematic graph is referred to as a {\em kinematic switch}, which is a type of {\em mode switch}.  
After the switch, as the manipulator moves, the poses of both the \pddl{Plate} and \pddl{Pizza} change with respect to the {\em world};  we move through this mode using the same actuators as before, but the feasible configuration space has changed. 

\subsubsection{Constrained motion planning}
\label{sec:constrained}

When the robot interacts with objects in the world, the effective configuration space is no longer the degrees of freedom of the robot:  it corresponds to the state of the {\em whole} system; we denote this space $\mathcal{W}$.   This state can be described by the discrete structure encoded in a kinematic graph, as well as continuous values of the transformations on the edges, which encode static relationships.
However, these systems are generally {\em under-actuated}, meaning that they cannot be locally controlled in arbitrary directions, because we can only directly actuate the robot's degrees of freedom.
Despite this, we can indirectly manipulate these objects by controlling the robot.

\begin{figure}[h]
    \includegraphics[width=0.75\textwidth]{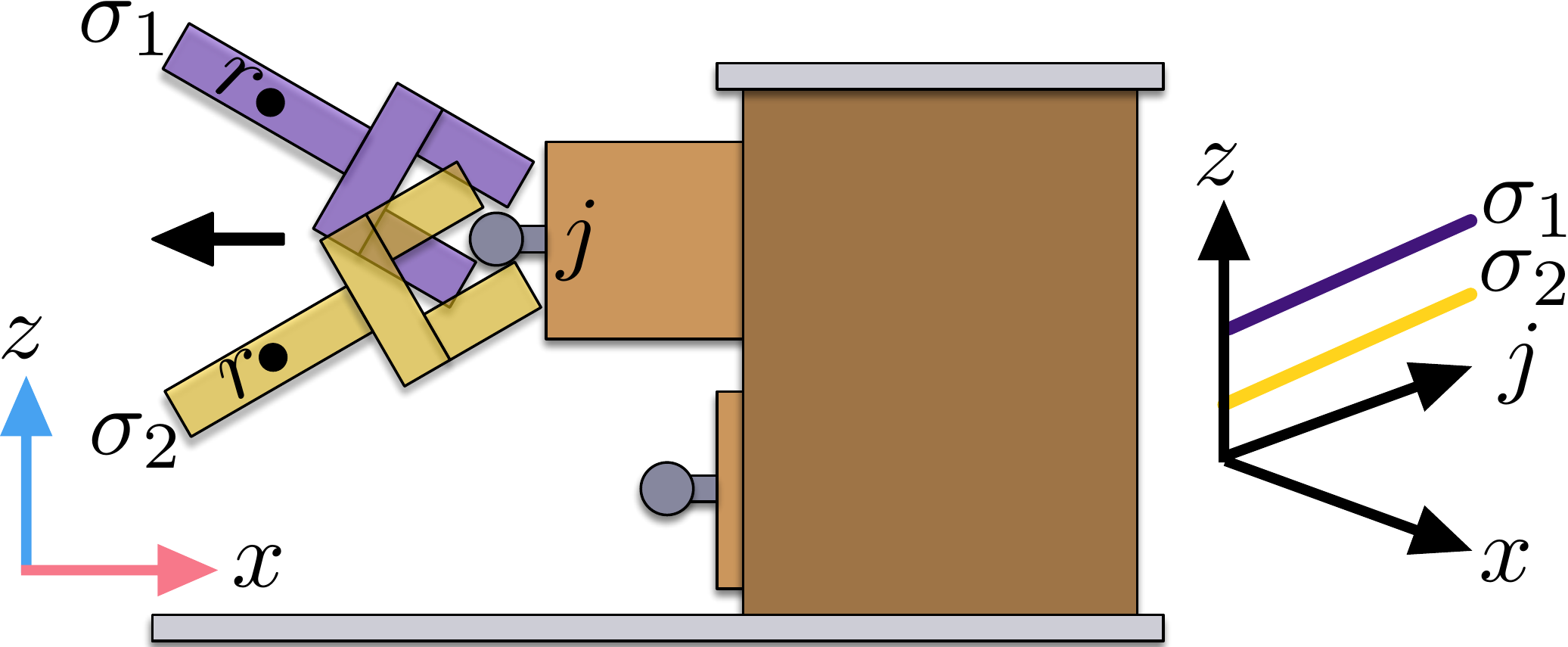}
    \caption{Constrained motion planning for a system in which a gripper pulls a drawer. The pose of the gripper relative to the drawer handle induces the 1D mode constraints $\sigma_1, \sigma_2$.}
    \label{fig:drawer}
\end{figure}

We begin by considering a simple ``single-mode'' problem in which the kinematic graph is fixed. \figref{fig:drawer} ({\em left}) illustrates a robot gripper pulling a drawer where the gripper pose is fixed relative to the drawer.  Although normally the gripper can translate generally in $x,z$, the drawer only has a single degree of freedom, denoted by joint $j$.
The combined configuration space of the gripper and the drawer is a three-dimensional space with coordinates $\langle  x, z, j \rangle$, but they are constrained by the need to keep the gripper attached to the drawer.

In this and many other manipulation problems, the constraint function $F(w)$, which now applies to the whole world configuration $w \in \mathcal{W}$, includes both the collision-free constraint and a kinematic constraint, which causes $\mathcal{W}_F$, the subspace of $\mathcal{W}$ for which $F$ holds, to be lower dimensional than $\mathcal{W}$.
As a result, sampling $\mathcal{W}$ randomly will have zero probability of producing a sample in $\mathcal{W}_F$, rendering standard sampling-based motion planning methods ineffective.
This difficulty of sampling motivated the development of {\em constrained motion planners}, which explicitly take these constraints into account and plan within the low-dimensional space $\mathcal{W}_F$.

Dimensionality-reducing constraints are often expressed using a mode parameter $\sigma$, a fixed value that affects the constraint $F_\sigma(w)$.  In general, $\sigma$ is real-valued. Here, we illustrate the effect of two different choices of this value, $\sigma_1$ and $\sigma_2$.  Each stipulates a different rigid attachment pose between the gripper and the drawer handle.
\figref{fig:drawer} ({\em right}) illustrates the combined configuration space and the feasible spaces $\mathcal{W}_{F_{\sigma_1}}$ and $\mathcal{W}_{F_{\sigma_2}}$, which are lines. 
The modes $\sigma_1$ and  $\sigma_2$ allow motion of the gripper along different 1D lines in this 3D space, depending on the grasp of the drawer handle.

The most general approach for constrained motion planning defines sampling and connecting operations that {\em project} values onto the constraint surface.  This is typically done by starting at a sampled point $\mathcal{W}$ and performing local descent on the constraint violation until convergence.
Because this is a numeric optimization, the constraint will generally never be exactly satisfied, but the samples can get $\epsilon$-close to the surface for any $\epsilon > 0$.
Several approaches have provided probabilistically complete methods for constrained motion planning using projection~\cite{stilman2007task,stilman2010global} and atlas-based techniques~\cite{berenson2009manipulation,berenson2010probabilistically,berenson2011task}.  
See \cite{kingston2018sampling,kingston2019exploring} for a comprehensive survey of these techniques.

When the kinematic graph is a tree, the planning problem is much easier.
The set of pairwise rigidity constraints specify all poses of objects relative either to the world frame (fixed objects) or to the robot (grasped objects).  
This collection of poses collectively constitutes a mode $\sigma$.
We can sample full configurations for the system that exactly satisfy these constraints by simply sampling the robot's degrees of freedom $q$, and performing forward kinematics to derive the full configuration $w$. 

\subsubsection{Multi-modal motion-planning}




Constrained motion planning provides a framework for reasoning about systems with many degrees of freedom, but few actuators.
However, it assumes that the constraints themselves remain constant, and therefore is not expressive enough to model multi-step manipulation problems in which the robot must make and break contact, changing the kinematic graph, and therefore the active constraints on its motions. 
In order to model such problems, we must allow the mode to undergo discrete changes~\cite{HauserLatombe,hauser2010randomized,HauserIJRR11,barry2013hierarchical}.
The state of the system is $s = \langle w, \sigma \rangle$, where we can think of $w \in \mathcal{W}$ as the collective configurations of the robot and all other objects or mechanisms in the environment and $\sigma$ as the additional mode information, indicating for example which objects are currently attached to which others.
The control theory community analyzes reachability for a similar class of hybrid systems, except that they typically address problems with a finite set of modes but more complex continuous-time dynamics~\cite{alur1995algorithmic, alur2000discrete, henzinger2000theory}.
For the most common cases of {\sc mmmp}, we can refactor this representation, so that $s = \langle q, K \rangle$, where $q \in \mathcal{Q}$ is the robot configuration and $K$ is a kinematic graph which contains the mode information and implies poses of all the bodies in the system, but we will make our general presentation in terms of $\langle w, \sigma \rangle$.

More formally, a {\sc mmmp} problem consists of a finite set $\{\modefam_1, \ldots, \modefam_m\}$ of {\em mode families}, each of which has a real-valued parameter vector $\theta$.   
Associated with each mode $\sigma = \modefam(\theta)$ is a constraint function $F_{\sigma}$ on full system configurations.  
At any given time, the system state $\langle w, \sigma \rangle$ is in a single mode $\sigma$, but whenever $w \in  F_{\sigma'}$, 
the system may execute a {\em mode switch}, typically represented by a change to the kinematic tree, into mode $\sigma'$. 
The goal of an {\sc mmmp} is typically a set of full system configurations $\mathcal{W}_*$, and a solution has the form $[\sigma_0, \tau_0, \sigma_1, \tau_1, \ldots, \sigma_k, \tau_k]$, where $s_0 = \langle w_0, \sigma_0 \rangle$ is the initial state of the system, $\tau_i$ is a trajectory in $F_{\sigma_i}$, $\tau_0(0) = w_0$, $\tau_i(0) = \tau_{i-1}(1)$ for $i \in \{1, ..., k\}$, and $\tau_k(1) \in \mathcal{W}_*$.

As an example, we model pick and place tasks in this framework.
Modes in which the robot is not grasping any objects are {\em transit modes}, and modes in which the robot is holding an object are {\em transfer modes}~\cite{AlamiTwoProbs,Alami90ISRR,simeon2004manipulation}.
For a robot with a single gripper, there is a {\em transit} mode family for free motion and a {\em transfer} mode family for each object that it can grasp.
In the transit mode family, the mode parameter is comprised of the fixed world poses of every movable object.
In the transfer mode family for a particular object, the mode parameter contains the grasp pose
as well as the fixed world poses of every other movable object.
Thus, although the system can only operate according to a single mode at a time, the mode parameter is high-dimensional because it contains constraints involving every movable object.
For interactions with cyclic kinematic graphs,
such as manipulating a drawer or opening a door, 
a constrained motion-planner (\sref{sec:constrained}) is generally required in order to plan within the mode. 


A key challenge in multi-modal motion planning is identifying configurations that are in the intersection of the constraint sets for two modes and thus allow the system to switch between them.  
This intersection is often lower dimensional than the feasible space $Q_{F_{\sigma}}$ of either mode.
In a pick-and-place domain, in order to perform a kinematic switch between a transit and transfer mode, the robot's gripper must be in contact with the involved object at a particular pose.
This requirement imposes 6 constraints on the robot's configuration, and as a result, the set of solutions is $(d-6)$-dimensional.  Fortunately, solutions can often be found using inverse kinematics ({\sc ik}) either by projecting random samples into the constraint set using optimization~\cite{wang1991combined} or by analytically solving for the solutions to a reparameterized set that captures its underlying dimensionality~\cite{diankov2010automated}. 

\begin{figure}[h]
    \includegraphics[width=0.95\textwidth]{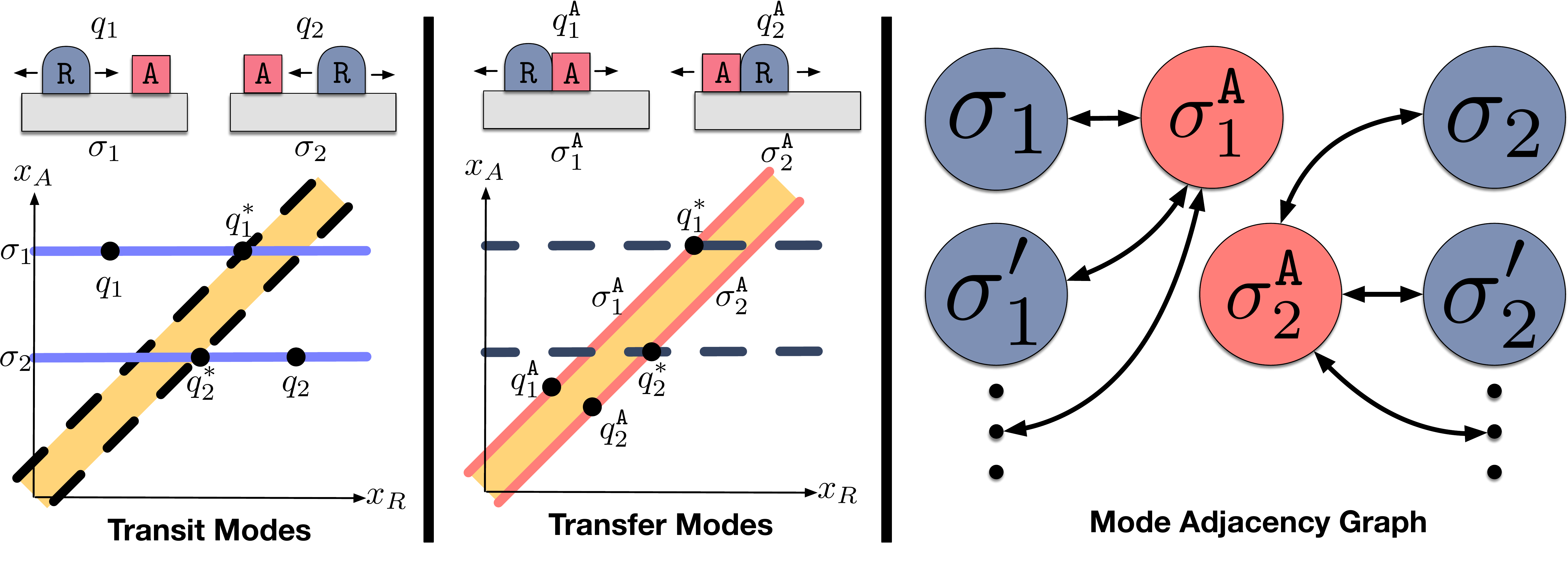}
    \caption{The feasible configuration space for two transit modes $\sigma_1, \sigma_2$ and two transfer modes $\sigma_1^\pddl{A}, \sigma_2^\pddl{A}$ modes. Mode switches between $\sigma_1 \leftrightarrow \sigma_2^\pddl{A}$ occur at configuration $q_1^*$. Mode switches between $\sigma_2 \leftrightarrow \sigma_2^\pddl{A}$ occur at configuration $q_2^*$.}
    \label{fig:transfer}
\end{figure}

\figref{fig:transfer} demonstrates a 1D robot (\pddl{R}) acting in the presence of a single movable object (\pddl{A}).
The two plots visualize the 2D combined configuration space of the robot and the movable object.
The left plot demonstrates the robot moving during a transit mode.
The two 1D blue lines indicate the space for which the system can change, which depends on the current mode $\sigma_1$ or $\sigma_2$.
These modes correspond to different placements of the movable object, which remains constant.
The yellow region corresponds to infeasible states where the robot and the object are in collision. 
Because the object can be placed anywhere on the interval, there are infinitely many possible transit modes.
The center plot demonstrates the robot and object moving during a transfer mode.
The robot can attach itself either to the left or right side of the object.
As a result, there are two possible transfer modes $\sigma_1^\pddl{A}, \sigma_2^\pddl{A}$, indicated by the 1D red lines.
The relative pose between the robot and object remains constant during a transfer mode.
The robot can switch between transit and transfer modes at a zero-dimensional (point) intersection between both lines ($q_1^*, q_2^*$).
The right figure visualizes legal mode transitions as a directed graph.
The transit modes $\{\sigma_1, \sigma_1', ...\}$ correspond to the robot being on the left of the object, whereas the transit modes $\{\sigma_2, \sigma_2', ...\}$ correspond to the robot being on the right of the object.
In order to switch to a new transit mode, the robot must first enter the appropriate transfer mode.
Finally, note that the graph is disconnected because the robot is unable to move to the other side of the object.

\subsection{Task planning}
\label{sec:task}

Within the AI community, there has been a long-standing focus on planning in discrete domains, generally with very large state spaces, but made tractable by using representations and algorithms that exploit underlying regularities in the structure of the domain.  
Ghallab {\it et al.}~\cite{ghallab2016automated} provide a comprehensive discussion of task planning from the AI perspective, and Karpas and Magazzeni~\cite{karpas2019automated} survey task planning for robotics.

The simplest formalization of AI planning is to specify a set of states (state space) ${\cal S}$, a set of transitions ${\cal T} \subseteq {\cal S} \times {\cal S}$ that describe legal changes to the state, an initial state $s_0 \in {\cal S}$, and a set of goal states $S_* \subseteq {\cal S}$.
Each directed transition $t = \langle s, s' \rangle \in {\cal T}$ moves the system from state $s$ to state $s'$.
The objective for a planner is to find a plan $\pi$, a sequence of transitions, that advances the initial state $s_0$ into a goal state $s_* \in S_*$.
This problem can be reduced to a graph traversal problem, where the vertices are states and directed edges are transitions, and solved using standard graph-search algorithms.  
However, the state spaces considered are very large, 
so it is critical to use a 
functional representation of ${\cal T}$ to ``reveal'' states incrementally, for example by working forward from the initial state. 


The first step toward compact representations and efficient algorithms is to {\em factor} the state representation into 
a collection of state variables.  
More formally, states can be represented using a set of variables ${\cal V} = \{1, ..., m\}$, each of which has a finite domain ${\cal X}_v$.
States are assignments of values $x_v \in {\cal X}_v$ for variables $v \in {\cal V}$.
This induces a state space ${\cal S} = {\cal X}_1 \times ... \times {\cal X}_m$ that is the Cartesian product of each variable's domain.
Consider a variation on the example in~\figref{fig:kinematic}, involving a single robot, a movable pizza, and a movable book. 
Each state specifies the locations of the robot, the pizza and the book,
and the object that the robot is holding (or \pddl{None}), as well as whether or not the pizza has been cooked.
The set of possible locations are: \pddl{Box}, \pddl{Plate}, \pddl{Table} and \pddl{Oven}. 
The robot can move between any pair of locations, pick up an object at the robot's current location if it is not holding anything, and place an object at the robot's current location.
We can describe a state as an assignment of values to the variables \pddl{atRob} (4 possible values), \pddl{at[Pizza]} (5 domain values), \pddl{at[Book]} (5 domain values), \pddl{holding} (3 domain values), and \pddl{cooked[Pizza]} (2 domain values). 
A state in this domain can then be defined as

\begin{small}
$$    \{\pddl{atRob}{=}\pddl{Plate}, \pddl{holding}{=}\pddl{None}, \pddl{at[Book]}{=}\pddl{Table},
    \pddl{at[Pizza]}{=}\pddl{Box}, \pddl{cooked[Pizza]}{=}\pddl{False}\}. $$
\end{small} 
Although in total the variables have 19 possible values, there are 600 possible states resulting from the possible combinations of variable values. 
Generally, the size of the state space grows exponentially in the number of variables.

Next, we need to encode the set of transitions compactly.
In many domains, due to locality of effect or other underlying domain properties, transitions change the value of only a small number of the state variables at a time, which allows us to describe large sets of transitions compactly using a single {\em action} that encodes the {\em difference} between the two states.
These changes can be described by a set of {\em effects} $\pddlkw{eff:}\; \{v_1 \gets c_1, ..., v_k \gets c_k\}$ that list the variables that are modified ($v_1, ..., v_k$) and their resulting values ($c_1, ..., c_k$).  This set of effects describes a large set of state-pairs in the transition:  one corresponding to each possible assignment to the values of the unchanged variables.

\begin{figure}[ht]
\begin{small}
\begin{lstlisting}
moveF[loc1,loc2]
 |{\bf pre}:| atRob|$\:=\:$|loc1
 |{\bf eff}:| atRob|$\:\gets\:$|loc2
moveH[loc1,loc2]
 |{\bf pre}:| atRob|$\:=\:$|loc1, holding|$\:=\:$|obj
 |{\bf eff}:| atRob|$\:\gets\:$|loc2
pick[obj,loc]
 |{\bf pre}:| atRob|$\:=\:$|loc, holding|$\:=\:$|None, at[obj]|$\:=\:$|loc
 |{\bf eff}:| at[obj]|$\:\gets\:$|None, holding|$\:\gets\:$|obj
place[obj,loc]
 |{\bf pre}:| atRob|$\:=\:$|loc, holding|$\:=\:$|obj
 |{\bf eff}:| at[obj]|$\:\gets\:$|loc, holding|$\:\gets\:$|None
cook[obj]
 |{\bf pre}:| at[obj]|$\:=\:$|Stove
 |{\bf eff}:| cooked[obj]|$\:\gets\:$|True
\end{lstlisting} 
\caption{A template specification of \pddl{moveF}, \pddl{moveH}, \pddl{pick}, \pddl{place}, and \pddl{cook} actions.
} \label{fig:sas-actions}
\end{small}
\end{figure}

Another structural property of many domains, which can be used to more compactly express legal transitions, is that each action may correctly be executed in only certain states.
For example, a \pddl{pick} action cannot be performed if the robot is already holding an object.
We can express this by specifying, for each action, a set of {\em preconditions} $\pddlkw{pre}:\; \{x_{v_1}{=}c_1, ..., x_{v_k}{=}c_k\}$ that describe the set of states in which that action can be executed in terms of values of some of the state variables.  

One final important structural property is an ``object-centric'' abstraction:  in most problems, the state variables correspond to properties of objects in the domain ({\it e.g.}, the location or color of a particular cup) or relations among them ({\it e.g.}, whether a particular cup is inside a particular box).  We can take describe the possible actions of a domain generically, via templates that are parameterized by a choice particular objects that are present in a domain instance.  This form of abstraction allows the size of the domain description to be independent of the {\em number} of state variables in the domain. 

We illustrate the basic principles of task planning via an example in 
\figref{fig:sas-actions}, which specifies the preconditions and effects of 
the \pddl{moveF}, \pddl{moveH}, \pddl{pick}, \pddl{place}, and \pddl{cook} actions.   This specification needs to be coupled with a listing of the {\em actual} entities in any domain instance, such as the names of objects (\pddl{Pizza} and \pddl{Book}) and locations (\pddl{Box}, \pddl{Plate}, \pddl{Oven}, and \pddl{Table}), to yield a complete transition-system specification. Then, the state variables are \pddl{holding}, \pddl{atRob}, along with \pddl{cooked[obj]} and \pddl{at[obj]} for each actual object name \pddl{obj}.   Similarly, the actual possible actions are generated by substituting all combinations of object and location constants in for the template variables. 
For example, with two objects and four locations, there are 8 instances of the \pddl{place} action.

To clarify the use of template variables, note the \pddl{pick} action description:  it is a {\em template} describing a finite number of action instances, one for each discrete value of \pddl{obj} and \pddl{loc}.  But note that these two variables play different roles.  As the number of possible values of \pddl{obj} increases, the {\em dimensionality} of the state of the problem (characterized by the {\em number} of state variables) increases;  as the number of possible values of \pddl{loc} increases, the {\em domains} of the \pddl{at[obj]} variables increases but the number of variables does not.


The final component of a planning problem is a description of the set of goal states, which has the same form as an action precondition, as a conjunction of values of some state variables, where all unmentioned state variables may have any arbitrary value.
For example, the following goal description encodes the entire set of states in which the pizza is cooked and on the plate: $\{\pddl{at[Pizza]}{=}\pddl{Plate}, \;\pddl{cooked[Pizza]}{=}\pddl{True}\}$.
The solution to a task-planning problem is a sequence of action instances $a_1, \ldots, a_k$, that induces a state sequence $s_0, \ldots, s_k$, where each $s_i$ is a state expressed as an assignment of values to state variables, $s_0$ is the initial state of the planning problem, $s_i$ satisfies the preconditions of $a_{i+1}$, $s_{i+1}$ is the result of executing $a_i$ in $s_i$, and $s_k$ satisfies the goal conditions.

To finish our example, let the initial state be
\begin{small}
$$s_0 = \{\pddl{holding}{=}\pddl{None}, \pddl{atRob}{=}\pddl{Table}, \pddl{at[Pizza]}{=}\pddl{Box}, \pddl{at[Book]}{=}\pddl{Table}, \pddl{cooked[Pizza]}{=}\pddl{False}\}.$$
\end{small}
Solving this task requires first placing the pizza on the oven to cook it and then relocating it to the plate:
\begin{align*}
    \pi = [&\pddl{moveF[Table,Box]}, \pddl{pick[Pizza]}, \pddl{moveH[Box,Oven]}, \pddl{place[Pizza]}, \\
    &\pddl{cook[Pizza]}, \pddl{pick[Pizza]}, \pddl{moveH[Oven,Plate]}, \pddl{place[Pizza]}].
\end{align*}

One focus of AI planning has been to define languages for specifying planning problems.
The one shown in \figref{fig:sas-actions} is similar to a lifted version of {\em simplified action specification} ({\sc sas}+)~\cite{backstrom1995complexity}.
The most widely-used formalism is {\em planning domain definition language} ({\sc pddl})~\cite{mcdermott1991regression}, which can be seen as a transition system where state variables are Boolean facts.
The AI planning community has developed {\em domain-independent} algorithms that can operate on any problem written in a planning language, without any additional information about the problem.  
A factored planning representation enables efficient algorithms for solving {\em relaxed problems}, simplified versions of the original problem, and using their solutions to estimate the distance to a goal state~\cite{helmert2006fast}.

Finally, there are several extensions to the basic task planning formalism~\cite{Fox03pddl2.1:an,edelkamp2004pddl2,fox2006modelling} that are relevant to {\sc tamp}.
One of these is {\em numeric planning}, which involves planning with real-valued variables such as time, fuel, or battery charge.
Recent approaches support planning with convex dynamics~\cite{bryce2015smt} and non-convex dynamics by discretizing time~\cite{coles2012colin}.  Although these methods have many use-cases, they 
currently cannot be directly applied to most {\sc tamp} problems because they assume the set of actions is finite.


\section{TASK AND MOTION PLANNING}
\label{sec:algorithms}
To find solutions to {\sc tamp} problems, we need to integrate aspects of motion planing, multi-modal motion planning, and task planning.  In this section, we introduce a framework for describing {\sc tamp} problems and algorithms that allows us to describe most of the broad range of existing methods within a unified framework, and which we hope elucidates modeling and algorithmic trade-offs among them.
We begin by providing a formalism for describing {\sc tamp} problems, 
then characterize solution methods in terms of their strategies for sequencing actions, for selecting their continuous parameters, and for integrating these methods. 


\subsection{TAMP problem description} \label{sec:problem}

Informally, {\sc tamp} problems use compact representational strategies from task planning to describe and extend a class of {\sc mmmp} problems.   
{\sc tamp} is an extension of {\sc mmmp} in that there may be additional state variables that are not geometric or kinematic, such as whether the lights are on or the pizza is cooked.
We begin by articulating a generic {\sc mmmp}, using an extension of a task-planning formulation, in \figref{fig:mmmp-as-tp}.  There are two extensions of the task-planning formalism visible here.  First, there are {\em continuous} action parameters.
Second, in addition to preconditions and effects we have a new type of clause, called \pddl{con} for {\em constraint}.  It is a set of constraints that all must hold true among the continuous parameters of the action in order for it to be a legal specification of a transition of the system. 

\begin{figure}[ht]
\begin{small} 
\begin{lstlisting}
moveWithin[i](|$\theta, w, \tau, w'$|)
 |{\bf con}:| $\tau(0) = w,\; \tau(1) = w',\; \big(\forall t \in [0, 1]\; F_{\modefam_\pddl{i}(\theta)}(\tau(t))\big)$
 |{\bf pre}:| mode|$\:= \modefam_\pddl{i}(\theta)\:$|, conf|$\:=w\:$|
 |{\bf eff}:| conf|$\:\gets\:w'\:$|
switchModes[i,j](|$w, \theta, \theta'$|)
 |{\bf con}:| $F_{\modefam_\pddl{i}(\theta_1)}(w),\; F_{\modefam_\pddl{j}(\theta_2)}(w)$
 |{\bf pre}:| mode|$\:= \modefam_\pddl{i}(\theta_1)\:$|, conf|$\:=w\:$|
 |{\bf eff}:| mode|$\:\gets \modefam_\pddl{j}(\theta_2)\:$|
\end{lstlisting}
\end{small}
\caption{A formalization of {\sc mmmp} in the style of task planning. There is a \pddl{moveWithin} action for each mode family $\Sigma_\pddl{i}$ and a \pddl{switchModes} action for each mode family pair $\Sigma_\pddl{i}, \Sigma_\pddl{j}$.}
\label{fig:mmmp-as-tp}
\end{figure}



This formulation does not extend to the basic formulation of {\sc mmmp}, but it provides a clear articulation of the overall system dynamics.   In a domain with a large number of objects, there will be a large number of mode families, each of which requires specifying a constraint on a very high-dimensional world configuration space.
What {\sc tamp} adds is the ability to ``unpack'' the entities in the problem description into sub-parts that are simpler to describe and that reveal substructure in the problem that enables algorithmic insights.  We will illustrate this process in a {\sc tamp} generalization of the ``cooking'' domain from \sref{sec:task} using one particular formalization style, shown in \figref{fig:tamp-actions}.  

Consider an example which has five movable objects, \pddl{A} through \pddl{E}.
We decompose the system configuration $w$ into state variables \pddl{atRob}, \pddl{holding}, \pddl{at[A]}, \pddl{at[B]}, \pddl{at[C]}, \pddl{at[D]}, and \pddl{at[E]}.
The discrete state variable \pddl{holding} can take values ranging over \pddl{\{None, A, B, C, D, E\}} and specifies the current mode family $\modefam$.
The variable \pddl{atRob} is now a robot configuration, and \pddl{at[obj]} is the pose of object \pddl{obj} relative to either the world coordinate frame (when $\pddl{holding} \neq \pddl{obj}$) or the robot hand coordinate frame (when $\pddl{holding} = \pddl{obj}$). 

The \pddl{moveF} (move while the gripper is free) and \pddl{moveH} (move while the gripper is holding) actions describe transit and transfer motion within modes.
The \pddl{pick} action corresponds to a switch from a transit mode to a transfer mode, while the \pddl{place} action corresponds to a switch from a transfer mode to a transit mode.

The sparsity of effect of planning action descriptions is a good match for articulating which state variables are changed (and, implicitly, which ones stay the same).  We can see that, in each action description, the \pddlkw{eff:} clause indicates just the variables that change.
When the preconditions involve discrete constant values (such as \pddl{None}), they are being used to specify the mode family of the initial state of the transition.
The advantage of being able to use templates is apparent:  the \pddl{moveH} action has a template parameter \pddl{obj}, meaning that there is a mode family for each object being held.


\begin{figure}[ht]
\begin{small} 
\begin{lstlisting}
moveF(|$q, \tau, q', p^\pddl{A}, \ldots, p^\pddl{E}$|)
 |{\bf con}:| Motion(|$q, \tau, q'$|), CFreeW(|$\tau$|), CFreeA(|$p^\pddl{A}, \tau$|), |\ldots|, CFreeE(|$p^\pddl{E}, \tau$|)
 |{\bf pre}:| holding|$\:=\:$|None, atRob|$\:=q\:$|, atA|$\:=p^\pddl{A}\:$|, |\ldots|, atE|$\:=p^\pddl{E}\:$|
 |{\bf eff}:| atRob|$\:\gets\:q'\:$|
moveH[obj](|$g, q, \tau, q', p^\pddl{A}, \ldots, p^\pddl{E}$|)
 |{\bf con}:| Motion(|$q, \tau, q'$|), CFreeW[obj](|$g, \tau$|),
      CFreeA[obj](|$p^\pddl{A}, g, \tau$|), |\ldots|, CFreeE[obj](|$p^\pddl{E}, g, \tau$|)
 |{\bf pre}:| holding|$\:=\:$|obj, at[obj]|$\:=g\:$|, atRob|$\:=q\:$|,  atA|$\:=p^\pddl{A}\:$|, |\ldots|, atE|$\:=p^\pddl{E}\:$|
 |{\bf eff}:| atRob|$\:\gets\:q'\:$|
pick[obj](|$q, p, g$|)
 |{\bf con}:| Stable[obj](|$p$|), Grasp[obj](|$g$|), Kin[obj](|$q, p, g$|)
 |{\bf pre}:| holding|$\:=\:$|None, atRob|$\:=q\:$|, at[obj]|$\:=p\:$|
 |{\bf eff}:| holding|$\:\gets\:$|obj, at[obj]|$\:\gets\:g\:$|
place[obj](|$q, p, g$|)
 |{\bf con}:| Stable[obj](|$p$|), Grasp[obj](|$g$|), Kin[obj](|$q, p, g$|)
 |{\bf pre}:| holding|$\:=\:$|obj, atRob|$\:=q\:$|, at[obj]|$\:=g\:$|
 |{\bf eff}:| holding|$\:\gets\:$|None, at[obj]|$\:\gets\:p\:$|
cook[obj](|$p$|)
 |{\bf con}:| Stable[obj](|$p$|), OnStove[obj](|$p$|)
 |{\bf pre}:| at[obj]|$\:=p\:$|
 |{\bf eff}:| cooked[obj]|$\:\gets\:$|True
\end{lstlisting}
\end{small}
\caption{One formalization of {\sc tamp} for an environment that contains the movable objects \pddl{A}, \pddl{B}, \pddl{C}, \pddl{D}, and \pddl{E}.
Actions now have real-valued parameters and constraints on these parameters. }
\label{fig:tamp-actions}
\end{figure}

Just as we have decomposed the configuration and the mode, we can decompose constraints, expressing them as conjunctions of constraints with smaller arity.  
For example, the \pddl{pick} action
has the constraint $\pddl{Kin[obj]}(q, p, g)$.  For any particular value of \pddl{obj}, representing an actual object in the domain, this represents a kinematic constraint, saying that if the robot is in configuration $q$ and holding object \pddl{obj} in grasp $g$, then \pddl{obj} will be at pose $p$.   
In \pddl{moveF} and \pddl{moveH}, the \pddl{Motion}($q, \tau, q'$) constraint specifies the relationship between a trajectory $\tau$ and two robot configurations, asserting that $\tau(0) = q$, $\tau(1) = q'$, and $\tau$ is continuous.
Notice that the trajectory $\tau$ appears neither in the preconditions nor effects of these actions; they are auxiliary parameters that describe motion {\em within} the modes.
The \pddl{Stable[obj]}$(p)$ constraint requires that $p$ be a pose representing a stable placement for object \pddl{obj} on a static object in the world.
Similarly, the \pddl{OnStove[obj]}$(p)$ constraint requires that $p$ be a stable placement where \pddl{obj} is specifically on a stove.
The \pddl{Grasp[obj]}$(g)$ constraint defines stable grasp poses (transforms between the hand frame and object frame) $g$ for object \pddl{obj}.
This set may be finite if there only a few known grasps but could be uncountably infinite, in general.
The collision-free constraint \pddl{CFreeA[obj]}$(p, g, \tau)$ asserts that if object \pddl{A} is at pose $p$, the robot is holding \pddl{obj} in grasp $g$, and it executes trajectory $\tau$, no collision will occur.
The constraint \pddl{CFreeW[obj]}$(g, \tau)$ is defined similarly except that it involves the fixed objects in the world (indicated by the abbreviation \pddl{W}).
Finally, although not pictured, because the $p^\pddl{A}, \ldots, p^\pddl{E}$ parameters in the \pddl{moveF} and \pddl{moveH} actions are each only mentioned in a single constraint and precondition, they can be compiled away using state constraints~\cite{lin1994state} or inference rules (axioms)~\cite{thiebaux2005defense}, resulting in these actions templates being independent of the number of objects in the problem instance.

\subsubsection{The form of solutions}

In preparation for studying algorithms for solving {\sc tamp} problems, it is useful to examine the form of a solution, which is a finite sequence of
action instances $\pi = [a_1, \ldots, a_k]$, where each $a_i$ includes assigned values for all parameters that satisfy that action's constraints. 
These actions induce a state sequence $[s_0, s_1, \ldots, s_k]$, where each $s_i$ is a state expressed as an assignment of values to state variables, $s_0$ is the initial state of the problem, $s_{i-1}$ satisfies the preconditions of $a_{i}$, $s_i$ is the result of executing $a_i$ in $s_{i-1}$, and $s_k$ satisfies the goal conditions.
Selecting the action templates and values for the template variables specifies the {\em form} of a solution, which we call a {\em plan skeleton}.  If the skeleton is fixed, the set of variables for which values must be selected is determined, and the problem that remains is one of selecting those values so that the constraints of the actions in the skeleton are satisfied. 

\begin{figure}[h]
    \includegraphics[width=\textwidth]{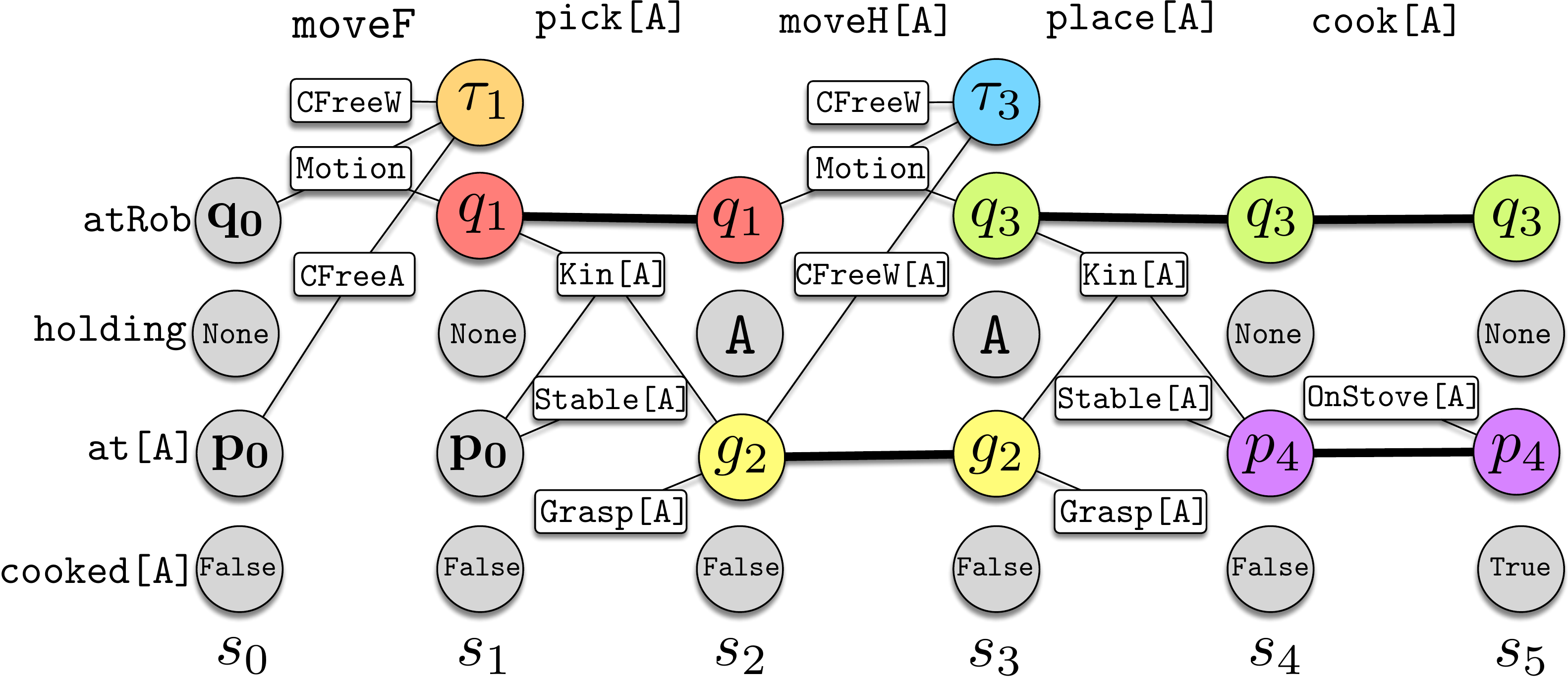}
    \caption{The plan skeleton from \eref{eqn:skeleton}. Grey values are constants. Thick black lines display equality constraints that persist over time.}
    \label{fig:factor}
\end{figure}

Consider a {\sc tamp} problem with a single movable object \pddl{A}.
Suppose the initial state is
$s_0 = \{\pddl{atRob}{=}\mathbf{q_0}, \pddl{at[A]}{=}\mathbf{p_0}, \pddl{holding}{=}\pddl{None}, \pddl{cooked[A]}{=}\pddl{False}\}$,
where the bold mathematical symbols $\mathbf{q_0}, \mathbf{p_0}$ are real-valued constants.
The set of goal states can be defined using conditions and constraints, such as $\pddl{cooked}{A}{=}\pddl{True}$.
One possible plan skeleton is:
\begin{align}  \label{eqn:skeleton}
    \pi = [&\pddl{moveF}(\mathbf{q_0}, \tau_1, q_1), \pddl{pick[A]}(q_1, \mathbf{p_0}, g_2), \\
    &\pddl{moveH[A]}(q_1, \tau_3, q_3), \pddl{place[A]}(q_3, p_4, g_2), \pddl{cook[A]}(p_4)]. \nonumber
\end{align}
where $q_1, \tau_1, g_2, q_2, \tau_3, p_4$ are the free parameters.
Plan skeletons can be visualized graphically by enumerating the sequence of $|\pi|+1$ values of each state variable, as well as motion parameters $\tau_1, \tau_3$, and associating the constraints of the $i$th action with the appropriate $i-1$ and $i$ state variables.
\figref{fig:factor} illustrates this plan skeleton in the form of a dynamic factor graph~\cite{dechter1992constraint}.
Round nodes represent state variables, and rectangular nodes represent constraints.  Gray nodes have constant values, and colored nodes represent variables.   Each vertical column corresponds to a state; the actions in the skeleton, which are responsible for the state changes are shown between the state columns, at the top.   Multiple nodes of the same color represent a single variable that is constrained to maintain its current value across multiple steps of the plan. Each constraint is connected to the variables it constrains.   Any assignment of values to the variables that satisfies all the constraints ``fills out'' the skeleton into a complete legal plan that is guaranteed to achieve the goal.  However, it may be the case that no satisfying assignment exists.

\subsection{Hybrid constraint satisfaction}  
\label{sec:hcsp}

\begin{figure}[h]
    \includegraphics[width=0.8\textwidth]{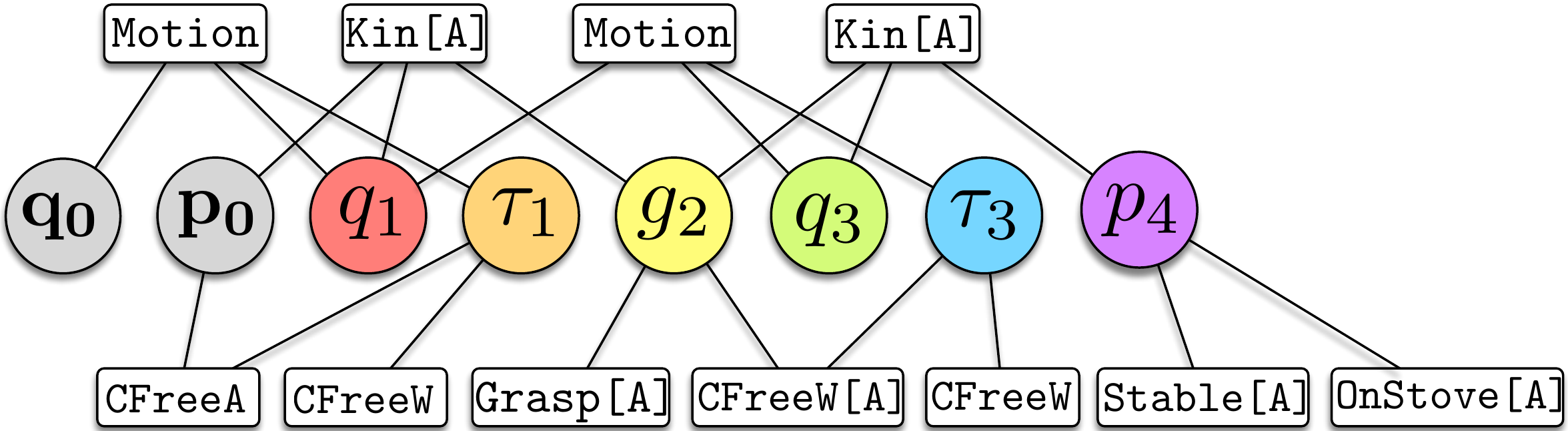}
    \caption{A simplification of the constraint network in \figref{fig:factor}.}
    \label{fig:network}
\end{figure}

Finding an assignment of values to the parameters of a plan skeleton that satisfy the associated constraints is a {\em hybrid constraint satisfaction problem} ({\sc h-csp}).
Although many parameters are inherently continuous, some may have discrete domains.
For example, there might be a finite set of stable resting surfaces for a particular object.
\figref{fig:network} compresses the plan skeleton in \figref{fig:factor} into a {\em constraint network}, a bipartite graph from parameters to constraints, by removing redundant constraints, constants, and parameters~\cite{dechter2003constraint}.
Although {\sc tamp} is decidable
via computational-geometry algorithms, just as in motion planning,  most practical approaches use optimization or sampling to solve the underlying {\sc h-csp}s.  Another dimension of variability in solution approaches is whether the method attempts to satisfy the entire constraint set at once or not:  methods vary dramatically in their high-level control structure for handling the search over skeletons and parameter values, and make different demands on constraint satisfaction methods.

\subsubsection{Joint Satisfaction} \label{sec:joint}

The most straightforward strategy for approaching an {\sc h-csp} is to reduce it to a constrained mathematical program and solve for values for all the free parameters at once.
Although there is a vast literature on mathematical programming, solving programs corresponding to {\sc tamp h-csp}s is often very difficult due to high dimensionality in continuous parameter space, the inclusion of discrete parameters, and the non-convexity of the constraints.   There is no efficient, general solution method for 
these mathematical programs.  
There are, nonetheless, some approaches of practical value.



When all decision variables are real-valued, a common solution strategy is to minimize an objective function, which incorporates both the {\em hard} constraint violation and any {\em soft} action cost penalties, using local-descent methods, though these are guaranteed to reach only a {\em local} optimum of the objective function,  which may not satisfy the constraints.
\eref{eqn:lgp} displays a mathematical program corresponding to the constraint network in \figref{fig:network}.
The trajectories $\tau_1$, $\tau_3$ are approximated as a sequence of robot configurations $\tau[0], \tau[1], ..., \tau[T]$ where $T$ is a hyperparameter.
Each constraint is associated with a real-valued (and often once or twice differentiable) function, which is expressed either in an equality ($g(\ldots)$) or inequality ($h(\ldots)$) constraint for the mathematical program.
Although it is not a focus of this survey, optimization can also fluidly incorporate action costs, enabling it to identify a solution that is not only feasible but also low-cost.
For example, \eref{eqn:lgp} minimizes the combined cost of moving through a \pddl{moveF} mode ($f_\pddl{moveF}(...)$) and a \pddl{moveH[A]} ($f_\pddl{moveH[A]}(...)$) mode, each of which are sums of a function defined on adjacent configurations that comprise trajectory parameter $\tau_1$ or $\tau_3$.
More generally, {\em mixed-integer programming} ({\sc mip}) techniques are required.
One prominent algorithm for solving {\sc mip}s is branch-and-bound, which performs a discrete search over assignments to the integer variables; then, 
conditioned on an assignment for each integer variable, the resulting mathematical program is real-valued and can be addressed by descent.  

\begin{equation} \label{eqn:lgp}
\begin{array}{rcll} 
\minimize\limits_{q_1, \tau_1, g_2, \tau_3, q_3, p_4   } &~& \sum_{t=1}^{T} f_\pddl{moveF}(   \tau_1[t], \tau_1[t-1]) + \sum_{t=1}^{T} f_\pddl{moveH[A]}(g_1, \tau_3[t], \tau_3[{}t-1])  & \\
\mathrm{subject~to} &~& g_\pddl{Grasp[A]}(g_1) = 0,\; g_\pddl{Stable[A]}(p_4) = 0,\; h_\pddl{OnStove[A]}(p_4) \leq 0 & \\ 
 &~& g_\pddl{Kin[A]}(q_1, \mathbf{p_0}, g_2) = 0,\; g_\pddl{Kin[A]}(q_3, p_4, g_1) = 0 & \\ 
 &~& h_\pddl{Motion}(\tau_1[t], \tau_1[t-1]) \leq 0,\; h_\pddl{Motion}(\tau_3[t], \tau_3[t-1]) \leq 0 & \text{for } t \in [T] \\ 
 &~& h_\pddl{CFreeW}(\tau_1[t]) \leq 0,\; h_\pddl{CFreeA}(\mathbf{p_0}, \tau_1[t]) \leq 0 & \text{for } t \in [T] \\ 
 &~& h_\pddl{CFreeW}(\tau_3[t]) \leq 0,\; h_\pddl{CFreeW[A]}(g_1, \tau_3[t]) \leq 0 & \text{for } t \in [T] \\ 
 &~& \tau_1[0] = \mathbf{q_0},\; \tau_1[T] = \tau_3[0] = q_1,\; \tau_3[T] = q_3 &
\end{array}
\end{equation}

\subsubsection{Individual Satisfaction} \label{sec:individual} 

An alternative approach to solving {\sc h-csp}s is to generate small groups of parameter values that satisfy a single constraint or a small set of constraints, and combine them. 
A {\em sampler} takes one or more constraints
and generates a sequence of assignments of values to the free parameters, where each assignment that is generated is guaranteed to satisfy the constraints.

A challenge when designing samplers is dealing with constraints whose set of satisfying values has lower dimension than combined domains of the free parameters.
For example, the $\pddl{Stable[obj]}(p)$ constraint requires object \pddl{obj} to rest perpendicular to a 2D plane within a 3D pose space, so this constraint lies in $\SE{2}$ despite the set of object poses being in $\SE{3}$.
The rejection-sampling strategy of sampling at random from a bounded region of $\SE{3}$ will have zero probability of producing a value satisfying this constraint.
However, samples can be produced by directly sampling \pddl{Stable} rather than $\SE{3}$.
Low-dimensional constraints remain problematic when attempting to produce values that also satisfy other constraints.
For example, consider solving for values of $q, p, g$ that satisfy both $\pddl{Stable[obj]}(p)$ and $\pddl{Kin[obj]}(q, p, g)$.
Here, the difficulty is finding a pose $p$ that satisfies \pddl{Stable} while also admitting values of $q, g$ that satisfy $\pddl{Kin}$.
One solution is to explicitly design samplers that operate on larger collections of constraints; this approach generally reduces to the joint satisfaction approach (\sref{sec:individual}).

Alternatively, one can design {\em conditional samplers} that take in {\em input} values for some of the parameters in the constraint(s) and produce satisfying {\em output} values for the rest of the parameters.
Intuitively, these samplers consume values already known to satisfy some constraints and find completing values that are compatible for additional constraints.
In the above example, a conditional sampler for \pddl{Kin[obj]} that takes in $p, g$ as inputs can consume a placement pose sampled by \pddl{Stable[obj]} and produce configurations $q$ through finding inverse kinematics ({\sc ik}) solutions.
In the event that no {\sc ik} solution exists, the conditional sampler returns an empty sequence, effectively rejecting the input values.
Boolean tests for a constraint can also be represented within this framework as degenerate ``samplers'' that perform a check on the input values but do not generate any output values.
For example, the collision-free constraints \pddl{CFreeW}, \pddl{CFreeA}, and \pddl{CFreeW[A]} can be evaluated by querying a collision checker.
In some applications, it may be beneficial to specify several conditional samplers for an individual constraint, which represent different partitions into input and output parameters.
For example, an alternative sampler for \pddl{Kin[obj]} takes in $q, g$ and performs forward kinematics to produce a pose for \pddl{obj} that satisfies the constraint.



\begin{figure}[h]
    \includegraphics[width=0.8\textwidth]{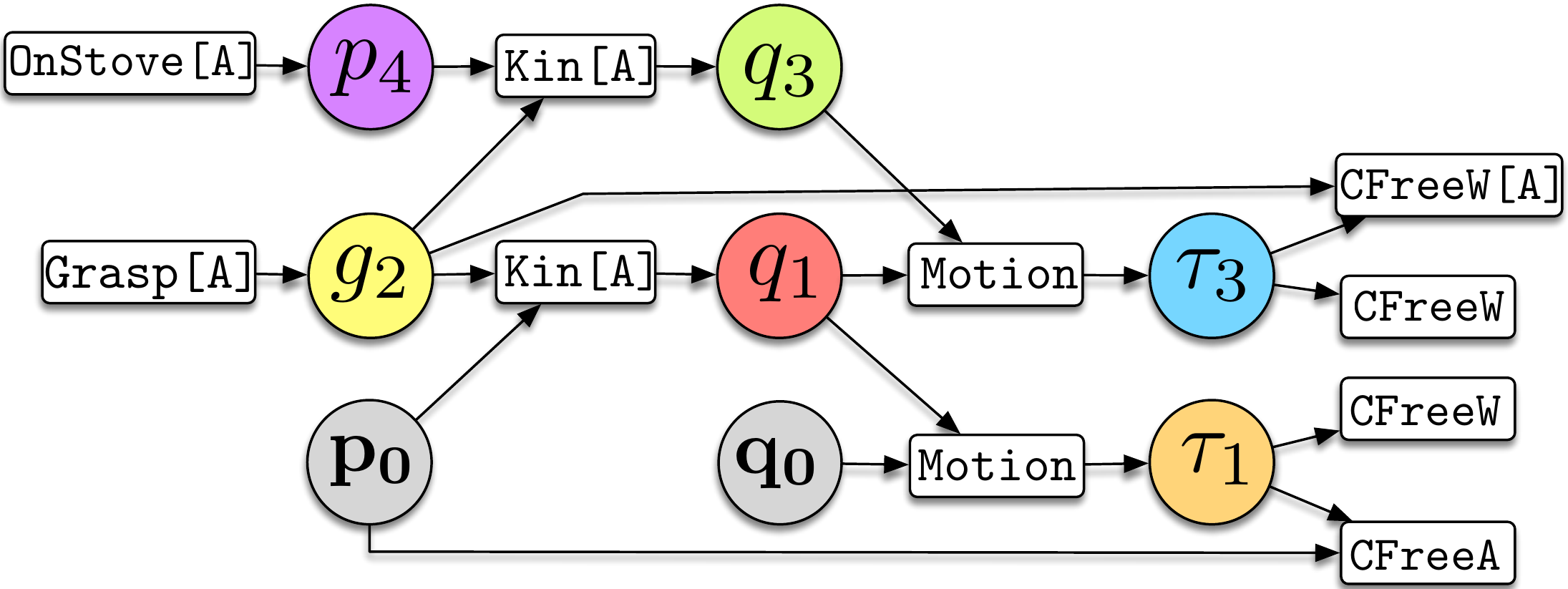}
    \caption{A sampling network for the constraint network in \figref{fig:network}.}
    \label{fig:oriented}
\end{figure}

More generally, several conditional samplers can be {\em composed} to form a {\em sampling network}~\cite{garrettIJRR2018}, a directed acyclic graph defined on free parameters and conditional samplers.
A directed edge from a parameter to a sampler indicates that the parameter is an input to the sampler.
A directed edge from a sampler to a parameter indicates that the parameter is an output of the sampler.
Each parameter is required to be the output of exactly one sampler.
This process is similar in spirit to converting a factor graph (constraint network) into a directed acyclic Bayesian network~\cite{dechter1992constraint}. 
\figref{fig:oriented} gives an example sampling network for the factor graph in \figref{fig:network}.

\subsubsection{Comparison} 

There is a trade-off between satisfying constraints individually versus jointly.
Individual satisfaction allows particular constraint types to be addressed using a special-purpose procedure, which is well equipped for that constraint, and provides a framework for modularly combining them.
For example, efficient algorithms for inverse kinematics and motion planning can be used to respectively generate robot configurations and trajectories.
Often, values generated in an attempt to satisfy one {\sc h-csp} can be reused in other, related {\sc h-csp}s.
In fact, values can even be usefully generated without a particular {\sc h-csp} in mind
as shown in \sref{sec:before}.


When jointly solving the complete set of constraints for a plan skeleton, only a single solution is required because, by construction, it has satisfied all relevant constraints.
In comparison, when constructing samplers and conditional samplers, it is important that they, in the limit as the number of samples goes to infinity, cover the complete space of feasible solutions, because some samples may be ruled out by other constraints in the problem.
Another advantage of joint satisfaction is that constraints on one parameter can transitively influence the selection of values for other parameters, directing the search.
Many methods for joint satisfaction requires the constraints to be made available in analytic form, enabling fast and accurate computation of derivatives used in descent methods.  
However, some constraints, such as collision constraints, are difficult to define in a differentiable form.  In such cases, sampling, which only requires black-box access to the constraint for use in rejection sampling, can be a more effective strategy, although its 
success is strongly dependent on the volume of solutions within the sampled space.

Finally, although we contrast these techniques, one can integrate both strategies.
For example, an algorithm could use individual sampling to generate values that satisfy \pddl{Stable[A]}, \pddl{Grasp[A]}, and \pddl{Kin[A]} but use joint satisfaction to solve for trajectories $\tau$ that satisfy \pddl{Motion}, \pddl{CFreeW}, \pddl{CFreeA}, and \pddl{CFreeW[A]}.

\subsection{Combining action sequence and parameter search} \label{sec:combining}

We now have the tools to search for action sequences (\sref{sec:task}) and to solve {\sc h-csp}s (\sref{sec:hcsp}).  In this section, we discuss strategies for combining them into integrated {\sc tamp} algorithms.  
We would like to order the decision-making in a way that minimizes the overall runtime of the algorithm, for a problem distribution. There are several intuitive principles for organizing the search, which are sometimes in conflict with one another.
One way to reduce search effort is to prune infeasible decision branches as quickly as possible (which is sometimes called failing fast~\cite{mandalika2019generalized}).  We would also prefer to postpone expensive computations until most of the rest of a potential solution is found.
For example, in many manipulation applications, collision checking is expensive, due to the geometric complexity of 3D meshes and the need to check at a fine resolution to ensure safety, so we might wish to {\em lazily} postpone this operation~\cite{bohlin2000path,dellin2016unifying}.
At the same time, we would like to balance the computational effort spent on each component,
for example, by not spending too much time trying to satisfy the {\sc h-csp} associated with a single skeleton or even single constraint, in case it is unsatisfiable.
Additionally, information gained in one branch of a high-level search, such as the solution or infeasibility of a subproblem, can often be re-used to make another branch of the search more efficient.

\begin{figure}[h]
    \includegraphics[width=0.75\textwidth]{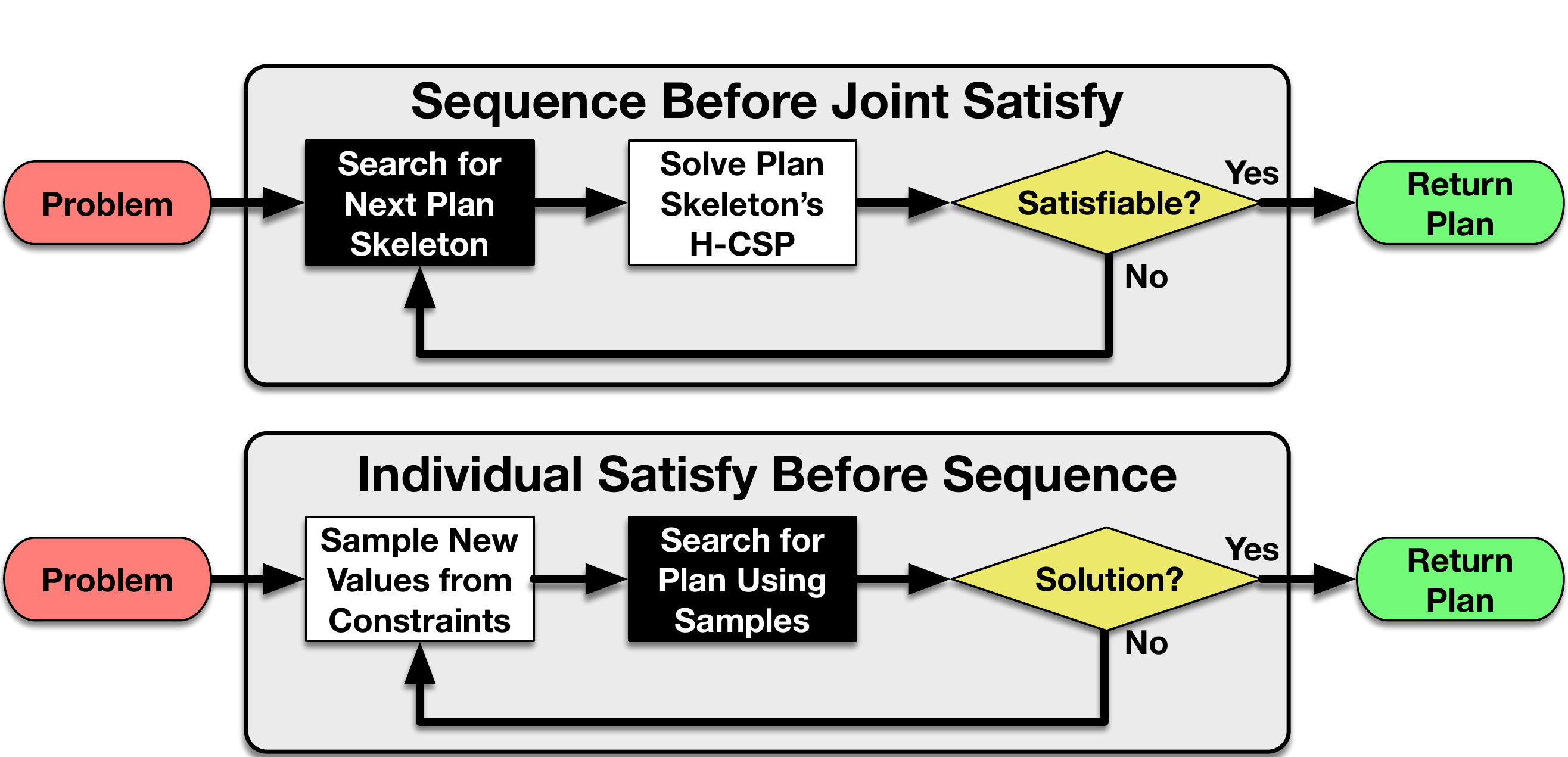}
    \caption{Flowcharts for two representative {\sc tamp} algorithms. {\em Top}: an algorithm that iteratively searches in the space of unbound plans and jointly satisfies the set of constraints (\sref{sec:sequencing}). 
    {\em Bottom}: an algorithm that iteratively performs individual sampling before searching in the space of fully-bound plans (\sref{sec:satisfaction}). 
    }
    \label{fig:representative}
\end{figure}

We begin by focusing on the overall control flow of {\sc tamp} algorithms, which determines the relative ordering of \action{}-sequencing and {\sc h-csp} (sub) problem solution.  There are three predominant classes of strategies:  {\it sequence before satisfy}, in which we find whole plan skeletons and then try to satisfy all of their constraints; {\it satisfy before sequence}, in which we find sets of satisfying assignments for individual constraints and attempt to assemble actions that use those values into complete plans; and {\em interleaved}, in which actions are added to the plan and additional constraints are satisfied incrementally.  We conclude by addressing an important aspect of making these approaches efficient, which is to take advantage of previous subproblem assignments or failures, in order to avoid re-addressing related subproblems.
\figref{fig:representative} illustrates the first two classes of {\sc tamp} strategies as flowcharts.

Throughout the discussion of control structures, it is important to remember that sampling and optimization techniques are typically only semi-complete, in that they are not able to certify that a problem instance is infeasible---they simply fail to find a solution in the time available to them.
Even for feasible {\sc h-csp}s these algorithms may still need to be run for an extremely long time if, for example, a feasible problem only admits a tiny volume of solutions.
Handling this is complicated by the fact that we might be forced to consider a possibly unbounded number of {\sc h-csp}s simultaneously.
To simplify the discussion, we can think about this process happening {\em non-deterministically}, where intuitively a separate thread is created for each {\sc h-csp}.
We can always simulate this behavior in a single process by appropriately revisiting the threads, making sure that none are starved.

\subsubsection{Sequencing first} \label{sec:sequencing}
The earliest algorithms for {\sc tamp} committed to a strict {\em hierarchy} of first finding an action sequence, and then finding continuous parameter values.
For example, Shakey~\cite{Nilsson84} performed {\sc strips} planning over high-level abstract actions, such as which room to move to, and then planned low-level motions that realized the high-level plan, with no mechanism for finding an alternative high-level plan if the lower-level motions were not possible.  So, Shakey aggressively assumed that problems satisfy the {\em downward refinement} property.  More formally, a two-level hierarchy satisfies the downward refinement property if {\em every} solution to the high level can be refined into a solution at the low level~\cite{bacchus1994downward}.
When this property holds, problems can be completely disentangled into separate task planning and motion planning problems, so an algorithm that determines a plan skeleton based on values of discrete template arguments strictly before solving the associated constraint-satisfaction problem is complete.
In {\sc tamp} problems in practice, downward refinement rarely holds.  As soon as geometric or kinematic considerations make some high-level plans infeasible (because, for example, three objects do not actually fit into the box we planned to put them in, or because the grasp needed to remove an object from a shelf will not work to place it on the stove and there is no surface available to use for regrasping), then we cannot inflexibly commit to any abstract plan without knowing its geometric and kinematic feasibility.

However, even when downward refinement does not hold, a top-down problem decomposition can be very effective, as long as there is a mechanism for ``backtracking'' and trying alternative high-level plans when the lower-level solver fails~\cite{srivastava2014combined,toussaint2015logic,garrettIJRR2018}.  
\figref{fig:representative} ({\em top}) illustrates this approach, in which there is an outer loop representing a search over legal plan skeletons;  for each plan skeleton, we attempt to solve the associated {\sc h-csp} and if we succeed, we return the complete solution, otherwise, we return to the outer loop and try another skeleton.
In many everyday {\sc tamp} applications, action sequencing is relatively inexpensive, making it advantageous to find a plausible action sequence before satisfying constraints.
Furthermore, solving {\sc h-csp}s can be computationally expensive, so by only attempting to solve {\sc h-csp}s that correspond to viable plan skeletons, we can potentially save substantial computation time.

\subsubsection{Satisfaction first} \label{sec:satisfaction}
\label{sec:before}
An alternative strategy is motivated by the fact that task planning in finite domains can often be very efficient in even very large problem instances, and therefore seeks to reduce the hybrid problem of {\sc tamp} to one or more discretized planning problems by generating values of continuous quantities, such as poses and configurations, and computing in advance which constraints they satisfy~\cite{simeon2004manipulation,HauserIJRR11,garrettIJRR2017,garrettIJRR2018}. 
For example, one might sample, for an environment with some fixed support surfaces, a set of values $p_i$ such that \pddl{Stable[A]}$(p_i)$ holds.
Approaches that perform satisfaction first almost always use individual satisfaction (\sref{sec:individual}), which is typically implemented using sampling, 
because they aim to generate values that are useful for a variety of plan skeletons. 
A single round of sampling will in general not suffice. When the discrete planning problem given a particular set of values is infeasible, it is necessary to generate more samples and try again, as illustrated in \figref{fig:representative} ({\em bottom}).   

Satisfying before sequencing is advantageous when the computational effort of repeatedly sequencing and failing to satisfy the associated {\sc h-csp} outweighs the computational effort of eagerly generating values that satisfy constraints upfront. 
This often is the case when one or more of the following are true: 1) sampling is efficient and does not result in a combinatorial explosion of sampled values, 2) each discrete action sequencing search has non-negligible overhead, and 3) sampled values are unlikely to satisfy critical constraints.



\subsubsection{Interleaved} \label{sec:interleaved}
There are many ways to interleave the searches for the action sequence and parameter values. 
In some cases, we would like to pre-sample state variable values, such as robot configurations and object poses, but defer the computation of motion parameter values, such as collision-free trajectories between two robot configurations.  In this case, the domains of the state variables have already been discretized. 
Conditioned on an assignment of values to every non-motion parameter (state variable) for an action instance, each motion parameter is only affected by the constraints of that particular action, which means that the problem of finding satisfying values for its parameters is independent of finding parameters for other actions.
Thus, the existence of a satisfying assignment to the motion parameters can be evaluated {\em online} during action sequencing in order to only compute values for action instances encountered during the search.
The strategy was first applied to {\sc tamp} under the name of semantic attachments~\cite{dornhege09icaps,dornhege09ssrr,dornhege2015task}. 
Although this strategy limits the amount of interleaving that is possible, it is appealing in that the state space is fixed during sequencing; it only identifies that some transitions are infeasible.

Interleaved action and parameter search can also aid the search for plan skeletons when sequencing first.
One source of search control is to observe that, in order for a plan skeleton to admit a satisfying assignment, all of its subsequences must also have satisfying assignments.
Thus, a partial plan skeleton can be pruned from the search if its {\sc h-csp} is infeasible. 
Some approaches even solve relaxations of the induced {\sc h-csp}s that omit certain constraints such as motion constraints with many decision variables, which are often satisfiable and thus are uninformative~\cite{LagriffoulDSK12,lagriffoul2014efficiently,lagriffoul2016combining,toussaint2015logic,toussaint2017multi, toussaint2018differentiable}.

A more general control structure is to perform a tree search, with layers alternating between selecting an action template and sampling parameter values for that action that satisfy the partial skeleton constraints.
A substantial difficulty in this approach is that tree nodes may have infinitely many successor nodes due to the possibly infinitely many satisfying parameter values and action instances that could be performed.
Thus, it is important that the search be {\em persistent}~\cite{GarrettIROS15,grey2016humanoid}, in the sense that it will revisit previous search nodes indefinitely in order to generate additional samples for continuous parameters.

\subsection{Communication between subproblems}
\label{sec:communication}
{\sc tamp} strategies require solving multiple {\sc h-csp} subproblems. 
These problems often have shared substructure 
that can be exploited, resulting in substantial reductions in computation. 
The primary algorithmic question is whether to share information about sets of constraints that {\em can} be satisfied ({\em positive}) or about sets of constraints that {\em cannot} be satisfied ({\em negative}).
Algorithms that satisfy constraints individually typically take the positive approach, 
and algorithms that satisfy constraints jointly typically take the negative approach.
Although we will discuss these approaches separately, it is possible to develop algorithms that use both, possibly to handle different types of constraints.

\subsubsection{Positive methods} \label{sec:positive}
Positive methods are straightforward:  whenever any {\sc h-csp} is solved, whether it contains one or many constraints, they add each constraint in the {\sc h-csp}, along with its satisfying assignment, to a database of {\em constraint elements}, known solutions to constraints.
For methods that satisfy before sequencing (\sref{sec:satisfaction}), this database is used to instantiate action instances before sequencing.
If action sequencing fails to find a solution, the feedback is that the current database is insufficient and more values must be sampled.
Some methods that sequence before satisfying (\sref{sec:sequencing}) also use positive feedback.
The {\em focused} algorithm presented in ~\cite{GarrettRSS17,garrettIJRR2018,garrett2020PDDLStream} plans using a mixed set of sampled values and free parameters (optimistic values), which represent values that are not yet available, but that might potentially be generated by a sampler. 

\subsubsection{Negative methods} \label{sec:negative}
Alternatively, instead of recording solutions to constraints, an algorithm could identify unsatisfiable {\em counterexample} {\sc h-csp}s.
Because any {\sc h-csp} that contains an unsatisfiable subproblem is itself unsatisfiable, any {\sc h-csp} that contains a recorded counterexample can be pruned;  this can, in turn, prevent action sequencing from exploring plan skeletons that are as-yet unexplored, but have the same failure case as a previously explored skeleton.  
Since most {\sc h-csp} solvers are only semi-decision procedures, they can never determine with certainty that a problem is infeasible.
One way to handle this problem is to assume unsatisfiability initially, but, as discussed in \sref{sec:combining}, allocate a thread to each {\sc h-csp} that continually searches for a solution in case one exists.
If one of these threads returns with a solution, the {\sc h-csp} is removed from the counterexample set.

A key algorithmic concern here is identifying informative counterexamples.
The smaller a counterexample is, the more {\sc h-csp}s and thus plan skeletons it can prune.
For example, consider the constraint network in \figref{fig:network} and suppose that placement $\mathbf{p_0}$ is on a tall shelf that the robot cannot reach.
If we could isolate constraint $\pddl{Kin[A]}(q_1, \mathbf{p_0}, g_2)$ as the bottleneck, rather than the full {\sc h-csp}, any plan skeleton that attempts to pick \pddl{A} at its initial placement will be pruned, informing the planner that \pddl{A} cannot be manipulated.

The problem of identifying small counterexamples can itself be time-consuming, requiring the original {\sc h-csp} to be decomposed into smaller {\sc h-csp}s, each of which is individually tested for unsatisfiability.
Several {\sc tamp} approaches have proposed heuristic methods for diagnosing failure and repairing the problem~\cite{Erdem,erdem2015integrating,Srivastava,srivastava2014combined}.
There are also several good domain-independent strategies.
For problems in which the state variables are pre-sampled (\ref{sec:interleaved}),  
the remaining {\sc h-csp} is often disconnected, and thus each unsatisfiable connected component can be independently added as a counterexample~\cite{dantam2016tmp,dantam2018incremental,dantam2018task}.
There are methods from the discrete {\sc sat} literature that, for unsatisfiable propositional formulas, identify {\em unsatisfiable cores}~\cite{liffiton2008algorithms}, small subsets of constraint that cause unsatisfiability.
These ideas can be extended to continuous mathematical programs, where real-valued constraint violation feedback can improve the efficiency of the search for counterexamples~\cite{shoukry2016scalable,shoukry2017smc,shoukry2018smc}.

\subsection{Taxonomy}

\tref{tab:matrix} illustrates a representative set of {\sc mmmp} and {\sc tamp} algorithms, categorized in terms of how they solve for continuous parameter values and how they combine searching for the mode-family or task-level structure of a plan with searching for continuous values. 
This table is meant to provide broad coverage, but is not exhaustive.
Each row lists one of three strategies for integrating constraint satisfaction and action sequencing: satisfaction first (\sref{sec:satisfaction}), interleaved satisfaction and sequencing (\sref{sec:interleaved}), and sequencing first (\sref{sec:sequencing}).
Each column lists one of three strategies for performing constraint satisfaction: assuming the state variables are pre-discretized and solving for motion parameters, individual sampling (\sref{sec:individual}), and joint optimization (\sref{sec:joint}).

\begin{table}[]
\centering
\resizebox{\textwidth}{!}{%

\begin{tabular}{|
>{\columncolor[HTML]{EFEFEF}}l |l|l|l|}
\hline
 &
  \cellcolor[HTML]{EFEFEF}\textbf{Pre-discretized} &
  \cellcolor[HTML]{EFEFEF}\textbf{Sampling} &
  \cellcolor[HTML]{EFEFEF}\textbf{Optimization} \\ \hline
  
\begin{tabular}[c]{@{}l@{}}\textbf{Satisfaction}\\ \textbf{First}\end{tabular} &
  Ferrer-Mestres${}^\ast$~\cite{ferrer2017combined,Ferrer-MestresPlanningPlanning} &
  \begin{tabular}[c]{@{}l@{}}Sim\'eon${}^\dagger$~\cite{simeon2004manipulation}\\ Hauser${}^\dagger$~\cite{HauserLatombe,hauser2010randomized,HauserIJRR11}\\ Garrett${}^\ast$~\cite{GarrettWAFR14,garrettIJRR2017}\\
  Krontiris${}^\dagger$~\cite{krontirisRSS2015,krontiris2016icra}\\ Akbari${}^\ast$~\cite{akbari2016task}\\ Vega-Brown${}^\dagger$~\cite{vega2016asymptotically} 
  \end{tabular} &
   \\ \hline
   
\textbf{Interleaved} &
  \begin{tabular}[c]{@{}l@{}}Dornhege${}^\ast$~\cite{dornhege09icaps,dornhege09ssrr,dornhege13irosws}\\
  Gaschler${}^\ast$~\cite{gaschler2013kvp,gaschler2015extending,gaschler2018kaboum}\\ Colledanchise${}^\ast$~\cite{colledanchise2019towards}\end{tabular} &
  \begin{tabular}[c]{@{}l@{}}Gravot${}^\ast$~\cite{gravot2005asymov,Cambon}\\ Stilman${}^\dagger$~\cite{stilman2005navigation,StilmanWAFR06,StilmanICRA07}\\ Plaku${}^\dagger$~\cite{Plaku}\\ Kaelbling${}^\ast$~\cite{kaelbling2011hierarchical,IJRRBel}\\ Barry${}^\dagger$~\cite{BarryISER12,barry2013hierarchical,barry2013manipulation}\\ Garrett${}^\ast$~\cite{GarrettIROS15,grey2016humanoid}\\
  Thomason${}^\ast$~\cite{thomason2019unified}\\ Kim${}^\ast$~\cite{kim2020learning, LIS260}\\ Kingston${}^\dagger$~\cite{kingstoninforming}\end{tabular} &
  Fernandez-Gonzalez${}^\ast$~\cite{Fernandez-Gonzalez2018Scottyactivity:Optimization} \\ \hline
  
\begin{tabular}[c]{@{}l@{}}\textbf{Sequence}\\ \textbf{First}\end{tabular} &
  \begin{tabular}[c]{@{}l@{}}Nilsson${}^\ast$~\cite{Nilsson84}\\ Erdem${}^\ast$~\cite{Erdem,erdem2015integrating}\\ Lagriffoul${}^\ast$~\cite{LagriffoulDSK12,lagriffoul2014efficiently,lagriffoul2016combining}\\ Pandey${}^\ast$~\cite{Pandey12,deSilva}\\ Lozano-P\'erez${}^\ast$~\cite{lozano2014constraint}\\ Dantam${}^\ast$~\cite{dantam2016tmp,dantam2018incremental,dantam2018task}\\ Lo${}^\ast$~\cite{lo2018petlon}\end{tabular} &
  \begin{tabular}[c]{@{}l@{}}Wolfe${}^\ast$~\cite{WolfeICAPS10}\\ Srivastava${}^\ast$~\cite{Srivastava,srivastava2014combined}\\ Garrett${}^\ast$~\cite{garrettIJRR2018,garrett2020PDDLStream}
  \end{tabular} &
  \begin{tabular}[c]{@{}l@{}}Toussaint${}^\ast$~\cite{toussaint2015logic,toussaint2017multi,toussaint2018differentiable}\\ Shoukry${}^\ast$~\cite{shoukry2016scalable,shoukry2017smc,shoukry2018smc}\\ Hadfield-Menell${}^\ast$~\cite{hadfield2016sequential}\end{tabular} \\ \hline
\end{tabular}%

}
\vspace{2mm}
\caption{A table that categorizes {\sc mmmp} and {\sc tamp} approaches, based on how they solve {\sc hc-sps} and how they integrate with constraint satisfaction with action sequencing. Approaches for {\sc mmmp} are designated with ${}^\dagger$, and approaches for {\sc tamp} are designated with ${}^\ast$. Each table cell is listed chronologically.}
\label{tab:matrix}
\end{table}

\section{EXTENSIONS}
\label{sec:extensions}
There are many ways to extend the basic {\sc tamp} problem class and associated algorithms; these are areas of current active research and future interest.

\subsection{Kinodynamic systems}
We have focused on domains with 
quasi-static dynamics (after the robot executes an action, the objects end in a stable state which persists until the robot's next action) and simple rigid-body kinematics.  
Extending {\sc tamp} to handle deformable objects and liquids as well as to full dynamics, such as throwing, are important directions.
Several {\sc tamp} approaches have already demonstrated the ability to plan for kinodynamic systems~\cite{Plaku,toussaint2017multi, toussaint2018differentiable}.

\subsection{State and action uncertainty}
A critical issue when acting in the real world is uncertainty.  In the presence of future-state uncertainty, a planning algorithm might need to take into account multiple possible outcomes of an action and ensure that there are actions it can take in response, to avoid unlikely but disastrous outcomes. 
More difficult, but pervasive, is uncertainty about the present state.  In this case, the problem can be treated as a ``belief-space'' planning problem, in which the planner reasons explicitly about the agent's state of information about the world and takes actions both to gain information and to drive the world into a desired belief state. 
Several approaches for deterministic observable {\sc tamp} have been extended to handle these challenges.~\cite{IJRRBel,hadfield2015modular,phiquepal2019combined,garrett2020online}

\subsection{Planning and learning}

A critical question to ask is where {\sc tamp} models come from.   Most work in {\sc tamp} assumes perfect observability, control actuation, and knowledge of the kinematics and shape of objects.  
Machine learning methods can help with the process of acquiring models in non-ideal domains as well as speeding computation. 
In particular, learning methods can improve {\sc tamp} in several ways:
\begin{itemize}
    \item {\bf Learning models.} Given a controller, whether acquired via learning or hand-built, the constraints that allow us to characterize successful executions for the {\sc tamp} planner may not be obvious, but they, too, can be learned from experience~\cite{wang2020learning}.
    \item {\bf Learning search guidance.} Classic task-planning algorithms derive domain-independent search heuristics from the action descriptions, but there are opportunities to automatically learn domain-dependent search heuristics~\cite{yoon2006learning,shen2020learning}, in the form of policies or value function estimates~\cite{yoon2008learning} or action-orderings as well~\cite{GarrettIJCAI16}.  Learning search guidance has been hugely influential in games like Go~\cite{silver2016mastering}.  In {\sc tamp} problems, it is more difficult because it is much less clear how to encode the state of the problem (object shapes and poses) in a way that affords generalization from current function approximation methods, and because the goal must be encoded into the prediction as well, but there is initial progress in this area~\cite{kim2020learning,driess2020deep,chitnis2016guided}. 
    \item {\bf Learning sampling guidance.} 
    Many {\sc tamp} planners use conditional samplers as part of their strategy for solving underlying {\sc h-csp}s.  Learning can make sampling much more effective, in two different ways, one in which the learning happens {\em during} a single search process and one in which the learning happens across problem instances.
    In the forward-search algorithms that interleave selection of action and parameters, we can derive inspiration from {\em Monte Carlo tree search} ({\sc mcts})~\cite{kocsis2006bandit,browne2012survey}, in which experience with trying to expand nodes in a branch of the tree is used to form local estimates of the likelihood that a solution lies along that branch.
    Sampling for continuous parameter values can itself be similarly guided, using techniques for optimistic global optimization~\cite{munos2014bandits, LIS260}.
    Samplers can also be learned from previous experience using generative models such as Generative adversarial networks ({\sc gan}s)~\cite{kim2018guiding}.
\end{itemize}



\begin{summary}[SUMMARY POINTS]
\begin{enumerate}
\item {\sc tamp} selects the sequence of high-level actions that the robot should take, the hybrid parameter values that determine how the action is performed, and the low-level motions that safely execute the action.
\item {\sc tamp} approaches build on research in motion planning, multi-modal motion planning, and task planning.
\item Many {\sc tamp} approaches can be seen as integrating a search over plan skeletons (partially specified plans) and the satisfaction of constraints over hybrid action parameters.
\item Existing approaches can be usefully categorized according to how they address and integrate these two types of decisions.
\end{enumerate}
\end{summary}

\begin{issues}[FUTURE ISSUES]
\begin{enumerate}
\item Further investigation is needed of strategies that combine sampling and optimization approaches to {\sc tamp}.
\item {\sc tamp} methods should be extended to plan in more realistic environments that, for example, involve deformable objects, time, dynamics, liquids and other agents.
\item Uncertainty is central to all real-world robot applications; future {\sc tamp} methods should consider both future-state and present-state uncertainty.
\item Incorporating learning-based methods into planning will enable planners to reason with learned action models, requiring less human-provided domain knowledge.
\end{enumerate}
\end{issues}


\section*{DISCLOSURE STATEMENT}
The authors are not aware of any affiliations, memberships, funding, or financial holdings that might be perceived as affecting the objectivity of this review. 


\section*{ACKNOWLEDGMENTS}

We gratefully acknowledge support from NSF grants 1523767 and 1723381; from AFOSR grant FA9550-17-1-0165; from ONR grant N00014-18-1-2847; from the Honda Research Institute; and from SUTD Temasek Laboratories.
Caelan Garrett, Rachel Holladay, Rohan Chitnis and Tom Silver are supported by NSF GRFP fellowships.
Any opinions, findings, and conclusions or recommendations expressed in this material are those of the authors and do not necessarily reflect the views of our sponsors.

\bibliographystyle{ar-style3}
\bibliography{references}

\end{document}